\documentclass{article}
\usepackage{algorithm}
\usepackage{algpseudocode}
\usepackage{enumitem}
\usepackage{hyperref}
\usepackage{url}
\usepackage{comment}
\usepackage{booktabs}
\usepackage{graphicx}
\usepackage{soul}
\usepackage{subfig}
\usepackage{comment}
\usepackage{xfrac}
\usepackage{tikz}
\usepackage{pgfplots, pgfplotstable}
\usetikzlibrary{positioning, fit, shapes.geometric, bending, decorations.text, matrix, calc}
\usetikzlibrary{pgfplots.groupplots}
\usepgfplotslibrary{colorbrewer}
\pgfplotsset{compat=1.15}

\pgfplotsset{
    /pgfplots/legend image code/.code={%
        \draw[mark repeat=2,mark phase=2,#1] 
            plot coordinates {
                (0cm,0cm) 
                (0.19cm,0cm)
                (0.38cm,0cm)%
            };
    },
}

\makeatletter
\let\save@mathaccent\mathaccent
\newcommand*\if@single[3]{%
  \setbox0\hbox{${\mathaccent"0362{#1}}^H$}%
  \setbox2\hbox{${\mathaccent"0362{\kern0pt#1}}^H$}%
  \ifdim\ht0=\ht2 #3\else #2\fi
  }
\newcommand*\rel@kern[1]{\kern#1\dimexpr\macc@kerna}
\newcommand*\widebar[1]{\@ifnextchar^{{\wide@bar{#1}{0}}}{\wide@bar{#1}{1}}}
\newcommand*\wide@bar[2]{\if@single{#1}{\wide@bar@{#1}{#2}{1}}{\wide@bar@{#1}{#2}{2}}}
\newcommand*\wide@bar@[3]{%
  \begingroup
  \def\mathaccent##1##2{%
    \let\mathaccent\save@mathaccent
    \if#32 \let\macc@nucleus\first@char \fi
    \setbox\z@\hbox{$\macc@style{\macc@nucleus}_{}$}%
    \setbox\tw@\hbox{$\macc@style{\macc@nucleus}{}_{}$}%
    \dimen@\wd\tw@
    \advance\dimen@-\wd\z@
    \divide\dimen@ 3
    \@tempdima\wd\tw@
    \advance\@tempdima-\scriptspace
    \divide\@tempdima 10
    \advance\dimen@-\@tempdima
    \ifdim\dimen@>\z@ \dimen@0pt\fi
    \rel@kern{0.6}\kern-\dimen@
    \if#31
      \overline{\rel@kern{-0.6}\kern\dimen@\macc@nucleus\rel@kern{0.4}\kern\dimen@}%
      \advance\dimen@0.4\dimexpr\macc@kerna
      \let\final@kern#2%
      \ifdim\dimen@<\z@ \let\final@kern1\fi
      \if\final@kern1 \kern-\dimen@\fi
    \else
      \overline{\rel@kern{-0.6}\kern\dimen@#1}%
    \fi
  }%
  \macc@depth\@ne
  \let\math@bgroup\@empty \let\math@egroup\macc@set@skewchar
  \mathsurround\z@ \frozen@everymath{\mathgroup\macc@group\relax}%
  \macc@set@skewchar\relax
  \let\mathaccentV\macc@nested@a
  \if#31
    \macc@nested@a\relax111{#1}%
  \else
    \def\gobble@till@marker##1\endmarker{}%
    \futurelet\first@char\gobble@till@marker#1\endmarker
    \ifcat\noexpand\first@char A\else
      \def\first@char{}%
    \fi
    \macc@nested@a\relax111{\first@char}%
  \fi
  \endgroup
}
\makeatother

\usepackage{sidecap}
\usepackage{wrapfig}

\usepackage[square,numbers,sort&compress]{natbib}
\def\legendscale{0.8}

\def\data{\mathcal{D}}

\definecolor{colorA}{HTML}{1B9E77}
\definecolor{colorB}{HTML}{D95F02}
\definecolor{colorC}{HTML}{7570B3}
\definecolor{colorD}{HTML}{E7298A}
\definecolor{colorE}{HTML}{66A61E}
\definecolor{colorF}{HTML}{E6AB02}
\definecolor{colorG}{HTML}{A6761D}
\definecolor{colorH}{HTML}{666666}

\definecolor{fuchsia}{rgb}{1.0, 0.0, 1.0}

\definecolor{editone}{HTML}{0000FF}
\definecolor{edittwo}{HTML}{FF0099}
\definecolor{editthree}{HTML}{FF6600}

\usepackage{iclr2024_conference,times}
\iclrfinalcopy
\setlist[itemize]{leftmargin=.5cm} 

\newcommand\blfootnote[1]{%
  \begingroup
  \renewcommand\thefootnote{}\footnote{#1}%
  \addtocounter{footnote}{-1}%
  \endgroup
}

\usepackage{amsmath,amsfonts,bm}

\def\eqref#1{equation~\ref{#1}}

\def\1{\bm{1}}

\def\vmu{{\bm{\mu}}}

\def\vs{{\bm{s}}}

\def\vw{{\bm{w}}}
\def\vx{{\bm{x}}}

\def\vz{{\bm{z}}}

\def\mI{{\bm{I}}}

\def\mSigma{{\bm{\Sigma}}}

\DeclareMathAlphabet{\mathsfit}{\encodingdefault}{\sfdefault}{m}{sl}
\SetMathAlphabet{\mathsfit}{bold}{\encodingdefault}{\sfdefault}{bx}{n}

\DeclareMathOperator*{\argmin}{arg\,min}

\usepackage{amsthm}

\newtheorem*{proposition*}{Proposition}

\newtheorem*{claim*}{Claim}
\theoremstyle{definition}

\newtheorem{definition}{Definition}
\newtheorem*{definition*}{Definition}

\newcommand{\sdata}{{\cal S}}

\definecolor{DarkGreen}{RGB}{158, 6, 151}

\def\hideNotes{0} 
\if\hideNotes0
    
    \newcommand\richb[1]{{\color{red}\sf{[richb: #1]}}}
    \newcommand\sa[1]
    {{\color{purple}\sf{[Sina: #1]}}}
    
\else
    \renewcommand{\tableofcontents}{}
    
    \newcommand\lolo[1]{}
    \newcommand\richb[1]{}
    \newcommand\josue[1]{}
\fi
\pgfplotsset{every x tick label/.append style={font=\tiny, yshift=0.5ex}}
\pgfplotsset{every y tick label/.append style={font=\tiny, xshift=0.5ex}}

\usepackage[utf8]{inputenc} 
\usepackage[T1]{fontenc}    
\usepackage{url}            
\usepackage{booktabs}       
\usepackage{amsfonts}       
\usepackage{nicefrac}       
\usepackage{microtype}      
\usepackage{xcolor}         

\hypersetup{
  citecolor=blue,
  colorlinks=true,
}

\title{Self-Improving Diffusion Models \\ with Synthetic Data}

\author{%
{\small%
Sina Alemohammad\footnotemark[1],{\,}\footnotemark[2] ~ Ahmed Imtiaz Humayun\footnotemark[1],~ Shruti Agarwal\footnotemark[2],~ John Collomosse\footnotemark[2],~ Richard Baraniuk\footnotemark[1]} \\
\small \footnotemark[1]{~~}Rice University, \small \footnotemark[2]{~} Adobe Research
}

\newcommand{\proposedMethod}{SIMS}

\begin{document}

\maketitle

\begin{abstract}
The artificial intelligence (AI) world is running out of real data for training increasingly large generative models, resulting in accelerating pressure to train on synthetic data.
Unfortunately, training new generative models with synthetic data from current or past generation models creates an {\em autophagous} (self-consuming) {\em loop} that degrades the quality and/or diversity of the synthetic data in what has been termed {\em model autophagy disorder} (MAD) and {\em model collapse}. 
Current thinking around model autophagy recommends that synthetic data is to be avoided for model training lest the system deteriorate into MADness.
In this paper, we take a different tack that treats synthetic data differently from real data.
Self-IMproving diffusion models with Synthetic data (\proposedMethod{}) is a new training concept for diffusion models that uses self-synthesized data to provide {\em negative guidance} during the generation process to steer a model's generative process away from the non-ideal synthetic data manifold and towards the real data distribution. 
We demonstrate that \proposedMethod{} is capable of {\em self-improvement}; it establishes new records based on the Fréchet inception distance (FID) metric for CIFAR-10 and ImageNet-64 generation and achieves competitive results on FFHQ-64 and ImageNet-512.
Moreover, \proposedMethod{} is, to the best of our knowledge, the first {\em prophylactic} generative AI algorithm that can be iteratively trained on self-generated synthetic data without going MAD.
As a bonus, \proposedMethod{} can adjust a diffusion model's synthetic data distribution to match any desired in-domain target distribution to help mitigate biases and ensure fairness.

\end{abstract}

\begin{figure}[h]%
    \centering
    \begin{minipage}{0.02\linewidth}
    \vspace{-0.8cm}
        \begin{minipage}{\linewidth}
        \rotatebox{90}{ \footnotesize Base model}
        \end{minipage}
        \begin{minipage}{\linewidth}
        \vspace{0.6cm}
        \rotatebox{90}{ \footnotesize MAD} 
        \end{minipage}
        \begin{minipage}{\linewidth}
        \vspace{0.8cm}
        \rotatebox{90}{ \footnotesize \proposedMethod{}}
        \end{minipage}
    \end{minipage}
    \begin{minipage}{0.75\linewidth}
       \centering
        \subfloat[\centering ]{{\includegraphics[width=\linewidth]{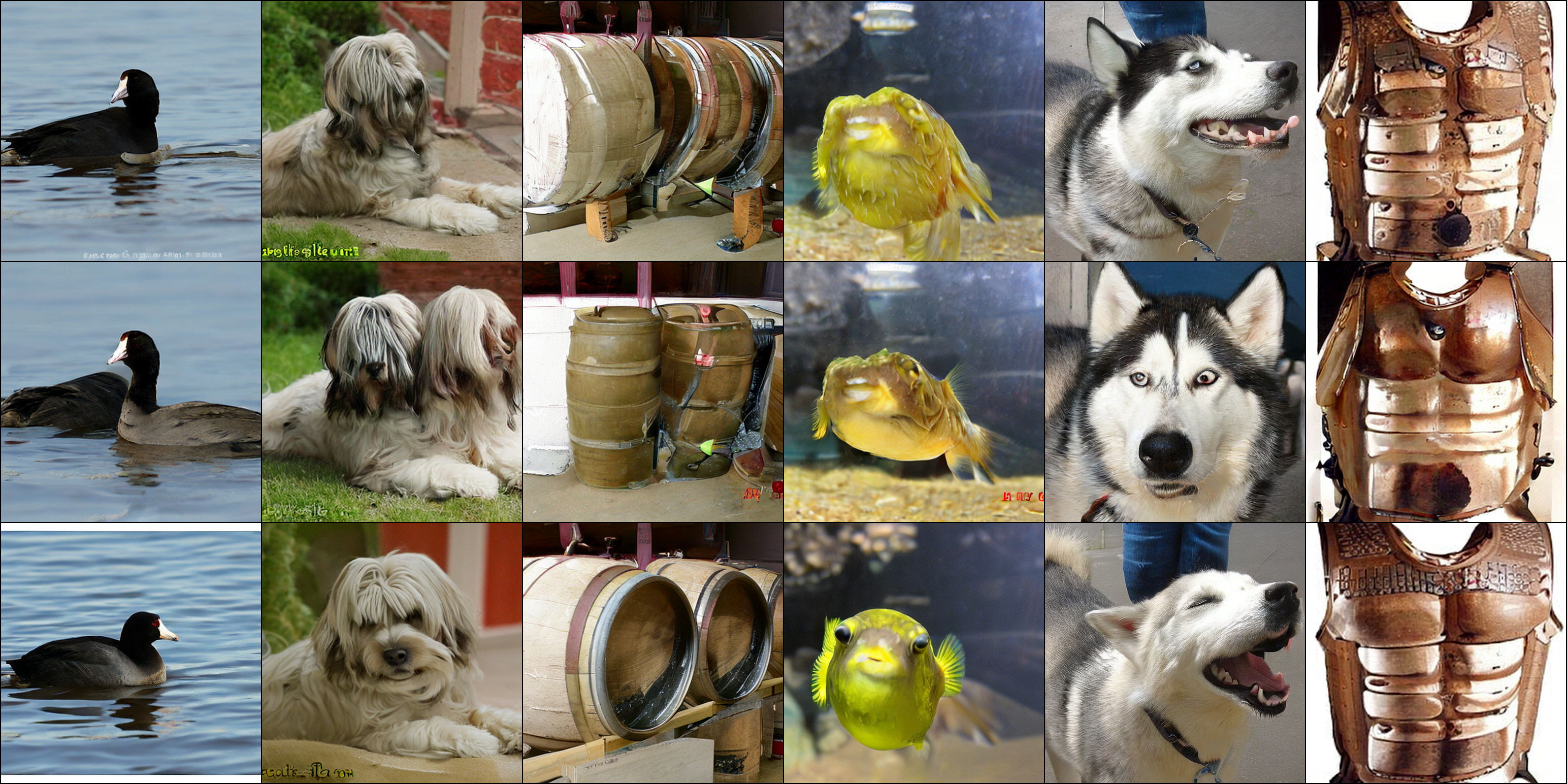} }}%
    \end{minipage}
    \caption{\small
    \textbf{Self-IMproving diffusion models with Synthetic data (\proposedMethod{})
    simultaneously improves diffusion modeling and synthesis performance while acting as a prophylactic against Model Autophagy Disorder (MAD)}.
    %
    First row: Samples from a base diffusion model (EDM2-S \citep{kynkaanniemi2024applying})    trained on $1.28$M real images from the ImageNet-512 dataset \cite{karras2024analyzing} (Fréchet inception distance, FID = $2.56$). 
    Second row: Samples from the base model after fine-tuning with $1.5$M images synthesized from the base model, which degrades synthesis performance and pushes the model towards MADness \citep{alemohammad2023arxiv, alemohammad2024selfconsuming} (FID = $6.07$). 
    Third row: Samples from the base model after applying \proposedMethod{} using the same  self-generated synthetic data as in the second row (FID = $1.73$).
    }
    \label{fig:mainfigure}%
\end{figure}
\section{Introduction}
\blfootnote{Corresponding author: 
sa86@rice.edu}Thanks to the ongoing rapid advances in the field of generative artificial intelligence (AI), 
we are witnessing a proliferation of synthetic data of various modalities that have been rapidly integrated into popular online platforms.
The voracious appetite of generative models for training data 
\citep{yahoomad,bigger_is_better, large_creative_ai, will_we_run_out} has caused practitioners to train new models either partially or completely using synthetic data from previous generations of models.
Synthetic training data is actually hard to avoid, because many of today's popular training datasets have been inadvertently polluted with synthetic data \citep{alemohammad2023arxiv, alemohammad2024selfconsuming}.

Unfortunately, there are hidden costs to synthetic data training.
Training new generative models with synthetic data from current or past generation models creates an {\em autophagous} (self-consuming) {\em loop} \citep{alemohammad2023arxiv, alemohammad2024selfconsuming} that can have a detrimental effect on performance.
In the limit over many generations of training, the {\em quality and/or diversity of the synthetic data will decrease}, in what has been termed Model Autophagy Disorder (MAD) \citep{alemohammad2023arxiv, alemohammad2024selfconsuming} and Model Collapse 
\citep{shumailov2023curse}.
MAD generative models also have major {\em fairness} issues, as they produce {\em increasingly biased samples} that lead to inaccurate representations across the attributes present in real data (e.g., related to demographic factors such as gender and race) \citep{wyllie2024fairness}.

MADness arises because synthetic data, regardless of how accurately it is modeled and generated, is still an approximation of samples from the real data distribution.\footnote{In this paper, by {\em real data} we mean direct samples from a target distribution. For example, in the context of natural images, real data would be digital photographs taken by a camera in a physical space. 
} 
An autophagous loop causes any  approximation errors to be compounded, ultimately resulting in performance deterioration and bias amplification.

Safely advancing the performance of generative AI systems in the synthetic data era requires that we make progress on both of the following open questions:
\begin{description}[style=unboxed, leftmargin=2em]

    \item[Q1.] How can we best exploit synthetic data in generative model training to improve real data modeling and synthesis?

    \item[Q2.] How can we exploit synthetic data in generative model training in a way that does not lead to MADness in the future?

\end{description}

In this paper, we develop {\em Self-IMproving diffusion models with Synthetic data} (\proposedMethod{}), a new learning framework for generative models that addresses both of the above issues simultaneously.
Our key insight is that, to most effectively exploit synthetic data in training a generative model, we need to change how we employ synthetic data. 
Instead of na\"{i}vely training a model on synthetic data as though it were real, \proposedMethod{} guides the model towards better performance but away from the patterns that arise from synthetic data training.

We focus here on SIMS for {\em diffusion models} in the context of image generation, because their robust guidance capabilities enable us to efficiently guide them away from their own generated synthetic data. 
In particular, we use a base model's own synthetic data to obtain a {\em synthetic score function} associated with the synthetic data manifold and use it to provide {\em negative guidance} during the generation process. 
By doing so, we steer the model's generative process away from the non-ideal synthetic data manifold and towards the real data distribution.

\begin{figure}[t]
    \centering
    \begin{tikzpicture}
    \draw (0,0) circle (1.5cm);
    \fill[green!10] (0,0) circle (1.5cm);
    \fill[black] (0,0) circle (2pt);
    

    \draw [->,thick] plot [smooth, tension=0.9] coordinates { (1.2cm, 0.9cm) (3.5cm, 1.47cm) (5.0cm, 1.7cm)};

    \draw [dashed, thick] plot [smooth, tension=0.9] coordinates { (0,0) (0.6cm,0.6cm) (1.2cm, 0.9cm) };
    
    \coordinate (intersection) at (1.2cm, 0.9cm); 

    \coordinate (aux) at (3.5cm, 1.47cm); 

    \draw[->,thick, purple, dashed] (intersection) -- (-2cm,0.2);

    \draw[<-,thick, purple] (3.4cm, 1.45cm) -- (intersection);
    
    \fill[blue] (intersection) circle (2pt);
    
    \fill[red] (aux) circle (2pt);


    \node[red, right] at (2.8cm,1.8cm) {\footnotesize $\vs_{\theta_{\rm s}}(\vx_t, t)$};


    \node[blue, right] at (0.3cm,1.2cm) {\footnotesize $\vs_{\theta_{\rm r}}(\vx_t, t)$};




    \node[purple,left] at (-2cm,0.0) {\footnotesize $-\omega \Delta \vs$};

    \node[purple,right] at (1.5cm, 0.5cm) {\tiny $\Delta \vs = \vs_{\theta_{\rm s}}(\vx_t, t) -\vs_{\theta_{\rm r}}(\vx_t, t)$};

    \node[right] at (5.0cm, 1.75cm) {\footnotesize Path to MADness};

    \node[black] at (0cm,-0.25cm) {\tiny Ground truth};

    \coordinate (sims) at (-0.16cm, 0.6cm); 
    \fill[colorE] (sims) circle (2pt);

    \node[colorE] at (-0.2cm, 0.9cm) {\footnotesize SIMS};
    \end{tikzpicture}
    \caption{\small\textbf{\proposedMethod{}
    simultaneously self-improves diffusion model modeling and synthesis performance while acting as a prophylactic against MAD.}
    \proposedMethod{} improves the score function $\vs_{\theta_{\rm r}}(\vx_t, t)$ for a base diffusion model trained on real data by training an auxiliary model on the same real data plus synthetic data from the base model.
    The score function $\vs_{\theta_{\rm s}}(\vx_t, t)$ of the auxiliary model can be combined with that of the base model to extrapolate a new score function (denoted \proposedMethod{}) that is closer to the real data distribution.
    }
    \label{fig:sims-trajectory}
\end{figure}
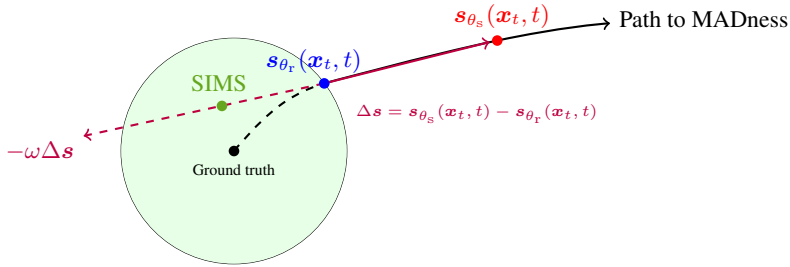

Figure~\ref{fig:sims-trajectory} depicts how \proposedMethod{} models and synthesizes more closely to the ground truth, real data distribution by {\em reversing the trajectory towards MADness}.
The green circle signifies the region in the function space of score functions that is inaccessible to a learning algorithm due to factors such as a limited amount of real data or sampling noise. 
As a result, training a first-generation base diffusion model on exclusively real data results in a score function $\vs_{\theta_{\rm r}}(\vx_t, t)$ (parameterized by a learnable neural network with parameters $\theta_{\rm r}$) in the vicinity of the ground truth.
Now, consider na\"{i}vely training a second-generation (auxilliary) model by fine-tuning the base model with synthetic data from the first-generation model. 
This corresponds to the synthetic augmentation loop in \citep{alemohammad2023arxiv, alemohammad2024selfconsuming}.
The resulting score function $\vs_{\theta_{\rm s}}(\vx_t, t)$ will be further away from the ground truth and on the path towards MADness.
Rather than tolerate this degraded second-generation model, we can use $\vs_{\theta_{\rm r}}(\vx_t, t)$ and $\vs_{\theta_{\rm s}}(\vx_t, t)$ to (linearly) extrapolate back into the inaccessible region.
Data generated using the resulting score function (denoted \proposedMethod{} in the figure) promise to be closer to the real data distribution than those from the base model.
Moreover, by explicitly reversing the MADness trajectory, diffusion models learned using SIMS promise to be less prone to MADness when the above process is repeated.

To summarize, given a training dataset, 
\proposedMethod{} performs the following four steps to obtain a self-improved diffusion model using self-generated synthetic data: 

\begin{algorithm}
\caption{\proposedMethod{} Procedure} \label{alg:sims}
\begin{algorithmic}[1]
    \Statex {\bf Input}: Training dataset $\mathcal{D}$
    \Statex {\bf Hyperparameters}: Synthetic dataset size $n_{\rm s}$, guidance strength $\omega$, training budget  $\mathcal{B}$

    \State \textbf{Train base diffusion model}: Use dataset $\data$ to train the diffusion model using standard training, resulting in the score function $\vs_{\theta_{\rm r}}(\vx_t, t)$.
    
    \State \textbf{Generate auxiliary synthetic data}:  Create an internal synthetic dataset $\sdata$ by generating $n_{\rm s}=|\sdata|$ samples from the base diffusion model.
    
    \State \textbf{Train auxiliary diffusion model}: Fine-tune the base model using only $\sdata$ within the training budget $\mathcal{B}$ to obtain $\vs_{\theta_{\rm s}}(\vx_t, t)$. Discard $\sdata$. 
    
    \State \textbf{Extrapolate the score function}: 
    Use $\vs_{\theta_{\rm s}}(\vx_t, t)$ to extrapolate backwards from $\vs_{\theta_{\rm r}}(\vx_t, t)$ to the \proposedMethod{} score function    
    \begin{equation*}
    \vs_{\theta}(\vx_t, t) = \vs_{\theta_{\rm r}}(\vx_t, t) - \omega (\vs_{\theta_{\rm s}}(\vx_t, t) -\vs_{\theta_{\rm r}}(\vx_t, t)) = (1 + \omega)\vs_{\theta_{\rm r}}(\vx_t, t) - \omega \vs_{\theta_{\rm s}}(\vx_t, t).
    \end{equation*}

    \Statex \textbf{Synthesize}:
    Generate synthetic data from the model using the \proposedMethod{} score function $\vs_{\theta}(\vx_t, t)$.
    
\end{algorithmic}
\end{algorithm}

We tally our contributions as follows:

\textbf{C1. Self-improvement}: We demonstrate that \proposedMethod{} makes significant progress on Q1 above by using self-generated synthetic data to significantly enhance the generation quality of image diffusion models.
\proposedMethod{} establishes new records based on the Fréchet inception distance (FID) metric \citep{fid} for CIFAR-10 and ImageNet-64 generation and achieves competitive results on FFHQ-64 and ImageNet-512 (see Table \ref{tab:comparison_sota} and Figure \ref{fig:self-improve}).
Moreover, given a diffusion model trained on a set of real data and a set of data synthesized from that model, we show that combining the real and synthetic data via \proposedMethod{} results in higher performance than either the original model or a model trained on the aggregate of the real and synthetic data (see Figures~\ref{fig:mainfigure} and~\ref{fig:eliminate_mad}).

\textbf{C2. MAD-prophylactic}: 
We demonstrate that \proposedMethod{} makes significant progress on Q2 above.
It is, to the best of our knowledge, {\em the first generative AI model that can be iteratively trained on self-generated, synthetic data without going MAD.}
We iterate the process in Algorithm~\ref{alg:sims} for 100 generations to show that there exist guidance parameter $\omega$ settings such that no MAD degradation occurs (see Figure \ref{fig:G-mad}).

\textbf{C3. Distribution controllability}: We show that \proposedMethod{} can adjust a diffusion model's synthetic data distribution to match any desired in-domain target distribution. This can help mitigate biases and ensure model fairness, all while improving the quality of the generated outputs (see Figure \ref{fig:dis_shift}).

Our findings clearly demonstrate that synthetic data can actually be both useful and safe for learning diffusion models and counters recent recommendations \citep{alemohammad2023arxiv, alemohammad2024selfconsuming,shumailov2023curse} that synthetic data is to be avoided in learning.
The difference in conclusions is due to \proposedMethod{}' unique approach: while training directly on (real data aggregated with) synthetic data causes a model to drift away from the true data distribution, \proposedMethod{} instead uses the synthetic data to explicitly avoid the synthetic data manifold and extrapolate closer to the true data distribution.

This paper is organized as follows.
Section~\ref{sec:back} overviews diffusion generative models, self-consuming loops, and past work attempting to slow or arrest MADness. 
Section~\ref{sec:3} presents the details of the \proposedMethod{} procedure that we outlined in Algorithm~\ref{alg:sims}.
Section~\ref{sec:results} exhibits the results of numerous computational experiments that demonstrated convincingly that \proposedMethod{} both improves model performance and either mitigates or completely prevents MADness.
We close with a discussion, recommendations, and direcions for future research in Section~\ref{sec:discussion}.

\section{Background}
\label{sec:back}

\textbf{Diffusion models.} Let $ p $ denote the distribution we seek to model. 
Diffusion models gradually diffuse the training data over time $ t \in [0, T] $ and sample from $ p $ by inversely modeling the forward diffusion process \citep{ho2020denoising, song2019generative}. 
Typically, this diffusion process involves transforming instances drawn from $ p $ into noisy versions with scale schedule $a_t$ and noise schedule $ \sigma_t $ at time $ t $. Hence, the conditional distribution of the noisy sample $\vx_t $ at time $ t $ can be formalized as
\begin{align}
    q_t(\vx_t|\vx_0) = \mathcal{N}(\vx_t \mid \vmu = a_t \vx_0, \mSigma = \sigma_t \mI),
    \label{noiseversion}
\end{align}
where $\vx_0$ is the data instance drawn from $ p $. 
The diffusion process can be formalized using a stochastic differential equation (SDE) \citep{song2019generative} 
\begin{align}
    d \vx = f(\vx,t) dt + g(t) d\vw,
    \label{forwardsde}
\end{align}
where $\vw$ is the standard Wiener process. 
Different choices for $f(\vx,t)$ and $g(t)$ result in different scaling $a_t$ and noise $\sigma_t$ schedules  in (\ref{noiseversion}). 
We refer the reader to \citep{karras2024analyzing} for more details on different SDE formulations for diffusion models. 

The solution to the SDE in (\ref{forwardsde}) is another SDE described by \citep{ANDERSON1982313}
\begin{align}
    d \vx = \Big[ f(\vx,t) - g^2(t) \nabla_{\vx_t} \log q_t(\vx_t) \Big] dt + g(t)d\Bar{\vw},
    \label{reversesde}
\end{align}
where $d\Bar{\vw}$ is the standard Wiener process when time flows in the reverse direction, and $q_t$ is the unconditional distribution in (\ref{noiseversion}) obtained by the forward SDE through (\ref{forwardsde}). The solution of the SDE in (\ref{reversesde}) starting from the samples of $\vx_T \sim q_T$ results in samples $\vx \sim q_0(\vx_0) $ that enable data generation from $p$. 

Since the score function $\nabla_{\vx_t} \log q_t(\vx_t)$ is unknown, the objective is to train a neural network with parameters $\theta$ to approximate the score function $\vs_\theta(\vx_t,t) \approx \nabla_{\vx_t} \log q_t(\vx_t)$ through
\begin{align}
     \min_{\theta}  \frac{1}{ | \mathcal{D} | }\sum_{\vx_0 \in \mathcal{D}} \mathbb{E}_{t \in [0,T], \vx_t \sim q_t(\vx_t|\vx_0)} \Big[\lambda(t) \| \vs_\theta(\vx_t,t) - \nabla_{\vx_t} \log q_t(\vx_t) \|^2 \Big],
     \label{trianmodel}
\end{align}
where $ \mathcal{D} $ is the training set containing samples from $p$, and $\lambda(t)$ is a temporal weighting function. 
The SDE in (\ref{reversesde}) can be solved by replacing $ \nabla_{\vx_t} \log q_t(\vx_t)$ with $\vs_\theta(\vx_t,t)$ and performing numerical integration. 
For conditional generation, one can also impose a condition on the score function during training to obtain the conditional score.

\textbf{Self-consuming generative models.} 
Let $\mathcal{A}(\cdot)$ represent an algorithm that, given a training dataset $\mathcal{D}$ as input, constructs a generative model with distribution $\mathcal{G}$, i.e., $\mathcal{G} = \mathcal{A}(\mathcal{D})$. 
Consider a sequence of generative models $\mathcal{G}^t = \mathcal{A}(\mathcal{D}^t)$ for $t\in \mathbb{N}$, where each model approximates some reference (typically real data) probability distribution $p_{\rm r}$.

\begin{definition}   
    \textbf{Self-consuming (autophagous) loop} \citep{alemohammad2023arxiv, alemohammad2024selfconsuming}: An autophagous loop is a sequence of distributions $(\mathcal{G}^t)_{t\in\mathbb{N}}$ where each generative model $\mathcal{G}^t$ is trained on data that includes samples from previous generation models $(\mathcal{G}^\tau)_{\tau=1}^{t-1}$.
\end{definition}

\begin{definition}
    \textbf{Model Authophagy Disorder (MAD)} \citep{alemohammad2023arxiv, alemohammad2024selfconsuming}: Let $\mathrm{dist}(\cdot, \cdot)$ denote a distance metric on distributions. A \emph{MAD generative process} is a sequence of distributions $(\mathcal{G}^t)_{t\in \mathbb{N}}$ such that 
    $\mathbb{E}[\mathrm{dist}(\mathcal{G}^{t} , p_{\rm r})]$ increases with $t$.
\end{definition}

One can form a variety of self-consuming loops based on how $\mathcal{D}^t$, the training data at generation $t$, is constructed from real data $\mathcal{D}_{\rm r}^t$ drawn from $p_{\rm r}$ and synthetic data $\mathcal{D}_{\rm s}^t$ generated by the model $\mathcal{G}^t$. 
Let the first generation model be trained solely on real data, i.e, $\mathcal{G}^1 = \mathcal{A}(\data_{\rm r})$. 
For subsequent generation models $\mathcal{G}^t = \mathcal{A}(\data^t), t \ge 2$, the three main loop types proposed in \citep{alemohammad2023arxiv, alemohammad2024selfconsuming} are based on how $\data^t$ is constructed:
\begin{itemize}

    \item {\bf Fully synthetic loop}: 
    Each model $\mathcal{G}^t$ for $t \geq 2$ trains exclusively on synthetic data sampled from models from the previous generation model, i.e., $\data^t = \data_{\rm s}^{t-1}$.
    
    \item {\bf Synthetic augmentation loop}:
    Each model $\mathcal{G}^t$ for $t \geq 2$ trains on the dataset $\data^t = \data_{\rm r} \cup \data_{\rm s}^{t-1}$ comprising a fixed set of real data $\data_{\rm r}$ from $p_{\rm r}$ plus synthetic data $\data_{\rm s}^{t-1}$ from the previous generation model.
    
    \item {\bf Fresh data loop}: Each model $\mathcal{G}^t$ for $t \geq 2$ trains on the dataset $\data^t = \data_{\rm r}^t \cup \data_{\rm s}^{t-1}$ comprising a fresh (new) set of real data $\data_{\rm r}^t$ drawn from $p_{\rm r}$ plus synthetic data $\data_{\rm s}^{t-1}$ from the previous generation model.
    
\end{itemize}
This paper focuses on the first two loop types above, which in general deteriorate into MADness of some kind.
In particular, for the fully synthetic loop, it has been shown theoretically and experimentally that $\mathbb{E}[\mathrm{dist}(\mathcal{G}^{\infty}, p_{\rm r})] \rightarrow \infty$ \citep{alemohammad2023arxiv, alemohammad2024selfconsuming}. 
In this scenario, often referred to as ``model collapse'' \citep{shumailov2023curse} in the literature, the sequence of models drifts away from the real data distribution until it no longer resembles it.  

\textbf{Mitigating MADness.} Several groups have developed methods to mitigate MADness, which we define as ensuring that $\mathbb{E}[\mathrm{dist}(\mathcal{G}^\infty, p_{\rm r})] \leq C$ for some bounded $C$. 
In words, the performance of a mitigated-MAD family of models does not diverge into full MADness ($C \rightarrow \infty$) but plateaus at a level that does not exceed the performance of the first-generation model, i.e., $\mathbb{E}[\mathrm{dist}(\mathcal{G}^\infty, p_{\rm r})] > \mathbb{E}[\mathrm{dist}(\mathcal{G}^1, p_{\rm r})]$.

\citep{bertrand2023stability, dohmatob2024tale} show that MADness can be mitigated in the synthetic augmentation loop.
The continuous inclusion of real data in the training set prevents the model from drifting too far from the initial model.
\citep{dohmatob2024model, gerstgrasser2024model} show that it is possible to mitigate MADness without incorporating real data in every generation, as long as the synthetic dataset size increases linearly across generations by accumulating synthetic data from all previous generations.

\textbf{Preventing MADness.} To more completely address the problem of performance degradation in self-consuming loops, one should aim to not just mitigate but \textit{prevent MADness}, where the sequence of model generations at least maintains and ideally improves on the performance of the first-generation base model, i.e., $\mathbb{E}[\mathrm{dist}(\mathcal{G}^\infty, p_{\rm r})] \leq \mathbb{E}[\mathrm{dist}(\mathcal{G}^1, p_{\rm r})]$.

The above results involve a closed loop, where the only external information about the target distribution $p_{\rm r}$ is a fixed initial real dataset. 
Incorporating new external information in self-consuming loops --- such as a verifier to oversee synthetic data selection \cite{feng2024beyond, setlur2024rl}, external guidance during the generation process \cite{gillman2024selfcorrecting}, or fresh real data \citep{alemohammad2023arxiv, alemohammad2024selfconsuming} --- has been shown to prevent MADness.

Research on self-consuming loops has not yet identified an approach where the inclusion of synthetic data in a closed loop with no external knowledge not only mitigates MADness across generations but completely prevents it.
In the next section, we introduce \proposedMethod{}, and in Section \ref{sec:MPresults}, we show that using \proposedMethod{} as the training algorithm $\mathcal{A}(\cdot)$ in the synthetic augmentation loop can fully prevent MADness.

\section{\proposedMethod: Self Improvement with Synthetic Data} \label{sec:3}

In this section, we develop the {\em Self-IMproving diffusion models with Synthetic data} (\proposedMethod{}) framework (recall Algorithm~\ref{alg:sims}) for improving the performance of a diffusion model using its own synthetic data; we term this {\em self-improvement}.
Note that while we explain \proposedMethod{} in the context of unconditional diffusion models, our method extends to conditional diffusion models as well.

\paragraph{\proposedMethod{}: Extrapolating to Self-Improvement.}
Let us unpack the \proposedMethod{} steps outlined in Algorithm~\ref{alg:sims} in the Introduction.
Consider a {\em base diffusion model} characterized by the score function $ \vs_{\theta_{\rm r}}(\vx_t, t) $ that was trained on real data samples drawn from a target real data distribution $p_{\rm r}$. 
At noise level $t$, the score function $\vs_{\theta_{\rm r}}(\vx_t, t)$ outputs a vector $\vz_r$ that points in the direction of increasing log probability density $\log p_{\rm r}$. 
By numerically solving the reverse SDE in (\ref{reversesde}) using the score function $\vs_{\theta_{\rm r}}(\vx_t, t)$, we obtain samples that follow the synthetic data distribution $p_s$. 
The goal is to have $p_s\approx p_{\rm r}$.
However, due to various factors --- including, but not limited to, the limited size of the training dataset, inaccuracies in solving the reverse SDE, and implicit algorithmic biases --- the synthetic data distribution $p_s$ does not exactly match the target distribution $p_{\rm r}$. This discrepancy results in a \textit{model-induced distribution shift.} 

To address this shift, we train a separate, {\em auxiliary diffusion model} using the same training hyperparameters used for the base model (i.e., for obtaining $\vs_{\theta_{\rm r}}(\vx_t, t)$) using a temporary, internal synthetic dataset $\sdata$ containing samples drawn from $p_s$. 
This results in the score function $\vs_{\theta_{\rm s}}(\vx_t, t)$. 
Since $\vs_{\theta_{\rm r}}(\vx_t, t)$ and $\vs_{\theta_{\rm s}}(\vx_t, t)$ are approximations of $p_{\rm r}$ and $p_s$, respectively, their difference serves as a useful surrogate for the model-induced distribution shift. 
By guiding the model away from $ \vs_{\theta_{\rm s}}(\vx_t, t) - \vs_{\theta_{\rm r}}(\vx_t, t) $ during the generation process, we can mollify
this shift. 
This motivates the {\em \proposedMethod{} score function} that we introduced in Algorithm~\ref{alg:sims} and reprise here:
\begin{equation}
    \vs_{\theta}(\vx_t, t) = \vs_{\theta_{\rm r}}(\vx_t, t) - \omega (\vs_{\theta_{\rm s}}(\vx_t, t) -\vs_{\theta_{\rm r}}(\vx_t, t)) = (1 + \omega)\vs_{\theta_{\rm r}}(\vx_t, t) - \omega \vs_{\theta_{\rm s}}(\vx_t, t),
\label{eq:sims}
\end{equation}
where $\omega$ is the guidance strength. 

The size of the synthetic dataset $n_{\rm s} = |\sdata|$ influences the effectiveness of \proposedMethod{}. 
As $n_{\rm s} \rightarrow \infty$, $\vs_{\theta_{\rm s}}(\vx_t, t) \rightarrow \vs_{\theta_{\rm r}}(\vx_t, t)$, and \proposedMethod{} collapses to the score function of the base model for any value of $\omega$, eliminating any room for guidance. 
Conversely, as $n_{\rm s} \rightarrow 0$ the estimate $\vs_{\theta_{\rm s}}(\vx_t, t)$ of $\vs_{\theta_{\rm r}}(\vx_t, t)$ will be poor, resulting in ineffective guidance. 
As a result, $n_{\rm s}$ is a hyperparameter for \proposedMethod{} that should be tuned for best performance. 
Our experiments indicate that selecting a synthetic dataset size that roughly matches the real dataset size works well in practice.

The training budget $\mathcal{B}$ of the auxiliary model has a more profound effect on \proposedMethod{}' performance. 
The goal is to obtain a score function $\vs_{\theta_{\rm s}}(\vx_t, t)$ that is neither too different from nor too similar to $\vs_{\theta_{\rm r}}(\vx_t, t)$. 
Therefore, we initialize $\vs_{\theta_{\rm r}}(\vx_t, t)$ with parameters $\theta_{\rm r}$ and then fine-tune with a training budget $\mathcal{B}$ on the synthetic dataset $\sdata$. 
In this paper, we quantify the training budget by how many images are seen by the model. 
When $\mathcal{B} = 0$ at the start of training, we have $\vs_{\theta_{\rm s}}(\vx_t, t) = \vs_{\theta_{\rm r}}(\vx_t, t)$, and \proposedMethod{} is equivalent to only using $\vs_{\theta_{\rm r}}(\vx_t, t)$ for data synthesis regardless of the value of $\omega$. 
As we increase $\mathcal{B}$, we can expect the \proposedMethod{} score function in (\ref{eq:sims}) to approach the ground truth distribution and then depart as the auxiliary model becomes influenced less by the training data $\data$ and more by the synthetic data $\sdata$.
Consequently, we stop the auxiliary model fine-tuning process at the optimizing $\mathcal{B}$.


\textbf{Inference Computational Cost.} \proposedMethod{} requires twice the number of function evaluations as the base model at inference time because of the auxiliary model (recall (\ref{eq:sims})). 
However, the number of function evaluations can be reduced with minimal impact on performance by applying guidance from the auxiliary model within a limited interval, as proposed in \citep{kynkaanniemi2024applying}, or by fine-tuning only a portion of the base model to obtain the auxiliary model. 
Appendix \ref{appendix:ablation} provides ablation studies regarding reducing the number of function evaluations and the effect of the synthetic dataset size on fine-tuning.

\textbf{Related Work.} Augmenting the score function of a diffusion model with guidance from external models has been an active research direction in diffusion-based generative modeling.
\cite{dhariwal2021diffusion} introduced the notion of classifier guidance, which involves training a separate conditional classifier using denoised images from a base model and using the gradients of the classifier to steer the denoising trajectory at every step. \cite{wallace2023end} introduced a method to perform plug-and-play guidance with pre-trained classifiers. \cite{ho2022classifier} introduced classifier-free guidance, where a diffusion model is trained to learn both a conditional and unconditional score function. 
During denoising, the unconditional score function is used as negative guidance, which leads to impressive gains in generation fidelity. \cite{kim2022refining} proposed discriminator guidance, where gradients of a discriminator are used to perform guidance. The discriminator is trained to classify between real images from the training dataset and synthetic samples generated by the target diffusion model. Discriminator guidance can be considered a proto \textit{self-improvement} method since it employs synthetic data from the base model to increase the realism of the generated samples. 
\cite{ahn2024self} proposed a method to self-improve conditional or unconditional diffusion models without any training --- by performing negative guidance using a clone of the base model with the attention weights replaced by an identity matrix, effectively resulting in a worse version of the base model.
A similar strategy was applied in the concurrent work of \cite{karras2024guiding}, who suggest using a ``bad'' version of the base model for negative guidance. The authors suggest training a reduced-parameter model for fewer training epochs to obtain a bad version of the base model. 
To draw an analogy, \proposedMethod{} can be interpreted as a method to create a bad version of the base model by training on its own synthetic data.

\begin{figure}[t]

\centering
\begin{tikzpicture}
    

\pgfplotsset{/pgfplots/group/every plot/.append style = {
    very thick
}};

    \begin{groupplot}[group style = {group size = 4 by 2, horizontal sep = 6mm}, width = 0.31\linewidth]

        \nextgroupplot[
        title ={\small CIFAR-10},
        ylabel={\small FID},
        xlabel={\small $w$},
        axis x line*=bottom,
        axis y line*=left,
        ymin=1.2,ymax=2.4,
        xmin=0,xmax=2,
        grid, legend style = {at={(0,0)}, nodes={scale=0.35, transform shape}, column sep = 0pt, legend to name = legend0, text=black, cells={align=left},}]

        \addlegendimage{empty legend}
        \addlegendentry{\hspace{-1.2cm}$\mathcal{B}$}

        \addplot[colorA, thick]
        table [
            x index=0,
            y index=1,
            col sep=comma]
            {csvfiles/mainfigure/cifar10_different_w_3.csv};
        
        \addplot[colorB, thick]
        table [
            x index=0,
            y index=2,
            col sep=comma]
            {csvfiles/mainfigure/cifar10_different_w_3.csv};
        
        \addplot[colorC, thick]
        table [
            x index=0,
            y index=3,
            col sep=comma]
            {csvfiles/mainfigure/cifar10_different_w_3.csv};

        \addplot[only marks, mark=*, mark size=.8pt, color=red] coordinates {(0.8,1.41)};
        \node at (axis cs:0.8,1.41) [anchor=north] {\tiny \textcolor{red}{1.41}};

        \addlegendentryexpanded{5 Mi};
        \addlegendentryexpanded{40 Mi};
        \addlegendentryexpanded{75 Mi};

    
    \nextgroupplot[
        title ={\small FFHQ-64},
        xlabel={\small $w$},
        axis x line*=bottom,
        axis y line*=left,
        ymin=0.5,ymax=3.5,
        xmin=0,xmax=3,
        grid, legend style = {at={(0,0)}, nodes={scale=0.35, transform shape}, column sep = 0pt, legend to name = legend1, text=black, cells={align=left},}]

        \addlegendimage{empty legend}
        \addlegendentry{\hspace{-1.2cm}$\mathcal{B}$}
        
        \addplot[colorA, thick]
        table [
            x index=0,
            y index=1,
            col sep=comma]
            {csvfiles/mainfigure/ffhq_different_w_2.csv};
        
        \addplot[colorB, thick]
        table [
            x index=0,
            y index=2,
            col sep=comma]
            {csvfiles/mainfigure/ffhq_different_w_2.csv};
        
        \addplot[colorC, thick]
        table [
            x index=0,
            y index=3,
            col sep=comma]
            {csvfiles/mainfigure/ffhq_different_w_2.csv};

         \addplot[only marks, mark=*, mark size=.8pt, color=red] coordinates {(1.5,1.04)};
        \node at (axis cs:1.5,1.04) [anchor=north] {\tiny \textcolor{red}{1.04}};

        \addlegendentryexpanded{4 Mi};
        \addlegendentryexpanded{34 Mi};
        \addlegendentryexpanded{110 Mi};

    \nextgroupplot[
        title ={\small ImageNet-64},
        xlabel={\small $w$},
        axis x line*=bottom,
        axis y line*=left,
        ymin=0.6,ymax=2,
        xmin=0,xmax=2,
        grid, legend style = {at={(0,0)}, nodes={scale=0.35, transform shape}, column sep = 0pt, legend to name = legend3, text=black, cells={align=left},}]

        \addlegendimage{empty legend}
        \addlegendentry{\hspace{-1.2cm}$\mathcal{B}$}
        
        \addplot[colorA, thick]
        table [
            x index=0,
            y index=1,
            col sep=comma]
            {csvfiles/mainfigure/imagnet64_100_2_different_w.csv};
        
        \addplot[colorB, thick]
        table [
            x index=0,
            y index=2,
            col sep=comma]
            {csvfiles/mainfigure/imagnet64_100_2_different_w.csv};
        
        \addplot[colorC, thick]
        table [
            x index=0,
            y index=3,
            col sep=comma]
            {csvfiles/mainfigure/imagnet64_100_2_different_w.csv};

         \addplot[only marks, mark=*, mark size=.8pt, color=red] coordinates {(0.9,0.92)};
        \node at (axis cs:0.9,0.92) [anchor=north] {\tiny \textcolor{red}{0.92}};

        \addlegendentryexpanded{31 Mi};
        \addlegendentryexpanded{56 Mi};
        \addlegendentryexpanded{73 Mi};

    \nextgroupplot[
        title ={\small ImageNet-512},
        xlabel={\small $w$},
        axis x line*=bottom,
        axis y line*=left,
        ymin=1.4,ymax=3,
        xmin=0,xmax=2,
        grid, legend style = {at={(0,0)}, nodes={scale=0.35, transform shape}, column sep = 0pt, legend to name = legend4, text=black, cells={align=left},}]

        \addlegendimage{empty legend}
        \addlegendentry{\hspace{-1.2cm}$\mathcal{B}$}
        
        \addplot[colorA, thick]
        table [
            x index=0,
            y index=1,
            col sep=comma]
            {csvfiles/mainfigure/imagnet512_100_2_different_w.csv};
        
        \addplot[colorB,  thick]
        table [
            x index=0,
            y index=2,
            col sep=comma]
            {csvfiles/mainfigure/imagnet512_100_2_different_w.csv};
        
        \addplot[colorC,  thick]
        table [
            x index=0,
            y index=3,
            col sep=comma]
            {csvfiles/mainfigure/imagnet512_100_2_different_w.csv};

        \addplot[only marks, mark=*, mark size=.8pt, color=red] coordinates {(0.7,1.73)};
        \node at (axis cs:0.7,1.73) [anchor=north] {\tiny \textcolor{red}{1.73}};

        \addlegendentryexpanded{50 Mi};
        \addlegendentryexpanded{102 Mi};
        \addlegendentryexpanded{200 Mi};

    \nextgroupplot[
        ylabel={\small FID},
        xlabel={\scriptsize $\mathcal{B}$ (Mi)},
        axis x line*=bottom,
        axis y line*=left,
        xmin=0,xmax=120,
        ymin=1.3,ymax=1.9,
        grid, legend style = {at={(0,0)}, nodes={scale=0.35, transform shape}, column sep = 0pt, legend to name = legendTCIFAR-10, text=black, cells={align=left},}]
        
        \addplot[colorB,thick]
        table [
            x index=0,
            y index=2,
            col sep=comma]
            {csvfiles/mainfigure/cifar10_different_training_2_all.csv};

        \addplot[colorC,thick]
        table [
            x index=0,
            y index=3,
            col sep=comma]
            {csvfiles/mainfigure/cifar10_different_training_2_all.csv};

        \addplot[colorA,thick]
        table [
            x index=0,
            y index=1,
            col sep=comma]
            {csvfiles/mainfigure/cifar10_different_training_2_all.csv};

        \addplot[only marks, mark=*, mark size=.8pt, color=red] coordinates {(40,1.41)};
        \node at (axis cs:41,1.41) [anchor=north] {\tiny \textcolor{red}{1.41}};

         \legend{$\omega = 0.6$, $ \omega = 0.9 $, $\argmin \omega$}

    \nextgroupplot[
        xlabel={\scriptsize $\mathcal{B}$ (Mi)},
        axis x line*=bottom,
        axis y line*=left,
        xmin=0,xmax=120,
        ymin=0.95,ymax=1.5,
        grid, legend style = {at={(0,0)}, nodes={scale=0.35, transform shape}, column sep = 0pt, legend to name = legendTffhq, text=black, cells={align=left},}]

        \addplot[colorB,thick]
        table [
            x index=0,
            y index=2,
            col sep=comma]
            {csvfiles/mainfigure/ffhq_different_training_2_all.csv};

        \addplot[colorC,thick]
        table [
            x index=0,
            y index=3,
            col sep=comma]
            {csvfiles/mainfigure/ffhq_different_training_2_all.csv};

         \addplot[colorA,thick]
        table [
            x index=0,
            y index=1,
            col sep=comma]
            {csvfiles/mainfigure/ffhq_different_training_2_all.csv};

        \addplot[only marks, mark=*, mark size=.8pt, color=red] coordinates {(34,1.04)};
        \node at (axis cs:34,1.04) [anchor=north] {\tiny \textcolor{red}{1.04}};

        \legend{$\omega = 1.1$, $ \omega = 1.5 $, $\argmin \omega$}

    \nextgroupplot[
        xlabel={\scriptsize $\mathcal{B}$ (Mi)},
        axis x line*=bottom,
        axis y line*=left,
        xmin =0, xmax = 200,
        ymin = 0.8, ymax = 1.4,
        grid,legend style = {at={(0,0)}, nodes={scale=0.35, transform shape}, column sep = 0pt, legend to name = legendT64, text=black, cells={align=left},}]

        \addplot[colorB,thick]
        table [
            x index=0,
            y index=2,
            col sep=comma]
            {csvfiles/mainfigure/imagnet64_different_training_all.csv};

        \addplot[colorC,thick]
        table [
            x index=0,
            y index=3,
            col sep=comma]
            {csvfiles/mainfigure/imagnet64_different_training_all.csv};

        \addplot[colorA,thick]
        table [
            x index=0,
            y index=1,
            col sep=comma]
            {csvfiles/mainfigure/imagnet64_different_training_all.csv};

        \addplot[only marks, mark=*, mark size=.8pt, color=red] coordinates {(56.6,.92)};
        \node at (axis cs:56.6,.92) [anchor=north] {\tiny \textcolor{red}{0.92}};

         \legend{$\omega = 0.6$, $ \omega = 1 $, $\argmin \omega$}

    \nextgroupplot[
        xlabel={\scriptsize $\mathcal{B}$ (Mi)},
        axis x line*=bottom,
        axis y line*=left,
        xmin =0, xmax = 200,
        ymin = 1.5, ymax = 2.6,
        grid,legend style = {at={(0,0)}, nodes={scale=0.35, transform shape}, column sep = 0pt, legend to name = legendT512, text=black, cells={align=left},}]

        \addplot[colorB,thick]
        table [
            x index=0,
            y index=2,
            col sep=comma]
            {csvfiles/mainfigure/imagnet512_100_2_different_training_all.csv};

        \addplot[colorC,thick]
        table [
            x index=0,
            y index=3,
            col sep=comma]
            {csvfiles/mainfigure/imagnet512_100_2_different_training_all.csv};

        \addplot[colorA,thick]
        table [
            x index=0,
            y index=1,
            col sep=comma]
            {csvfiles/mainfigure/imagnet512_100_2_different_training_all.csv};

        \addplot[only marks, mark=*, mark size=.8pt, color=red] coordinates {(102,1.73)};
        \node at (axis cs:102,1.73) [anchor=north] {\tiny \textcolor{red}{1.73}};

         \legend{$\omega = 0.6$, $ \omega = 0.9 $, $\argmin \omega$}

\end{groupplot}
\node at ($(group c1r1) + (-15pt,20pt)$) {\pgfplotslegendfromname{legend0}}; 
\node at ($(group c2r1) + (-15pt,20pt)$) {\pgfplotslegendfromname{legend1}}; 
\node at ($(group c3r1) + (-15pt,20pt)$) {\pgfplotslegendfromname{legend3}}; 
\node at ($(group c4r1) + (-15pt,20pt)$) {\pgfplotslegendfromname{legend4}}; 
\node at ($(group c2r2) + (20pt,20pt)$) {\pgfplotslegendfromname{legendTffhq}}; 
\node at ($(group c1r2) + (20pt,20pt)$) {\pgfplotslegendfromname{legendTCIFAR-10}}; 
\node at ($(group c4r2) + (20pt,20pt)$) {\pgfplotslegendfromname{legendT512}}; 
\node at ($(group c3r2) + (20pt,-20pt)$) {\pgfplotslegendfromname{legendT64}}; 

\end{tikzpicture}

\caption{\small \textbf{\proposedMethod{} consistently self-improves diffusion models.} Top row: FID between the \proposedMethod{} model from Algorithm~\ref{alg:sims} and the real data distribution as a function of the guidance parameter $\omega$ at three different checkpoints of the training budget $\mathcal{B}$ as measured by the number of million-images-seen (Mi) during fine tuning of the auxiliary model.
Bottom row: FID of the \proposedMethod{} model as a function of training budget 
for three different values of the guidance parameter $\omega$.
} 
\label{fig:self-improve}

\end{figure}
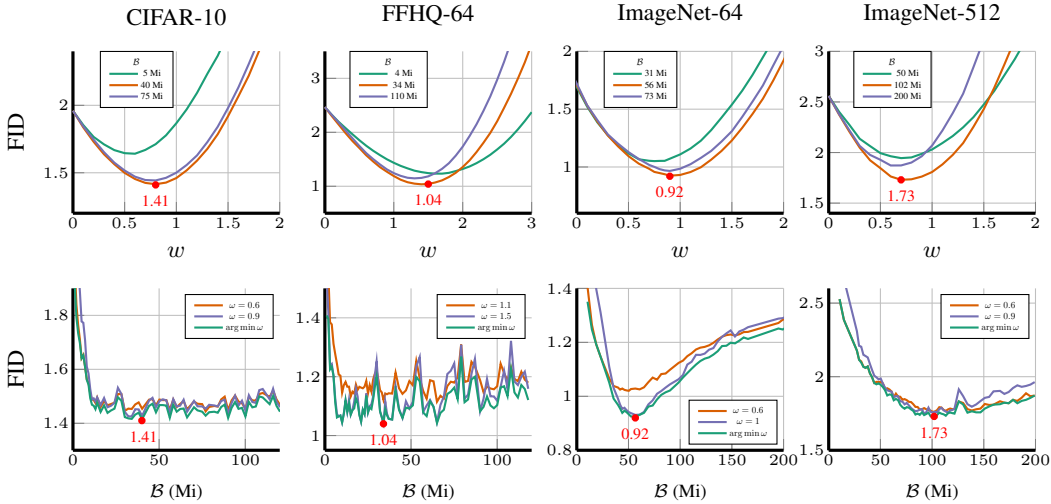

\section{Experimental Results}
\label{sec:results}

In this section, we present the results of an array of computational experiments with \proposedMethod{}.
In Section~\ref{sec:SIresults} we demonstrate that \proposedMethod{} makes significant progress on open question Q1 from the Introduction by self-improving the modeling performance of large-scale diffusion models using self-synthesized data.
In Section~\ref{sec:MPresults} we demonstrate that \proposedMethod{} makes significant progress on open question Q2 from the Introduction by acting as a prophylactic against MADness.
In Section~\ref{sec:DSresults} we show how \proposedMethod{} can adjust a diffusion model's synthetic data distribution to match any desired in-domain target distribution to mitigate biases and ensure model fairness.

\subsection{Self-Improving Diffusion Models} \label{sec:SIresults}

\textbf{Experimental Setup.} We use four diverse real image datasets $\data_{\rm r}$ for performance evaluation: $32\times 32$ resolution CIFAR-10 (50k images) \citep{krizhevsky2009learning}, $64\times 64$ resolution FFHQ-64 (70k images) \citep{karras2019style}, $64\times 64$ resolution ImageNet-64 (1.2M images), and $512\times 512$ resolution ImageNet-512 (1.2M images) \citep{deng2009ImageNet}. 

For CIFAR-10 and FFHQ-64, we use the unconditional Variance Preserving (VP) variant of the EDM diffusion model from \citep{karras2022elucidating} as the base model for \proposedMethod. 
For ImageNet-64 and ImageNet-512, we use the conditional EDM2-S model from \citep{karras2024analyzing}. 
While we use RGB-space diffusion models for CIFAR-10, FFHQ-64, and ImageNet-64, the ImageNet-512 model operates as a latent diffusion model with a latent space dimensionality of $64\times64\times4$. 
For all experiments with ImageNet-512,  we keep the encoder-decoder VAE fixed and use StabilityVAE \citep{rombach2022high} as in \citep{karras2024analyzing}. 
For all models, we use Heun's second-order solver \citep{Süli_Mayers_2003} for the de-noising process as proposed in \citep{karras2022elucidating}. 

For each base model, we use the publicly available code and pre-trained model weights from \citep{karras2022elucidating, karras2024analyzing}. 
To train each auxiliary model (recall Algorithm~\ref{alg:sims}), we first generate $n_{\rm s}=|\sdata|$ synthetic data samples from the base model and then fine-tune the base model using $\sdata$ and the same training configuration as the base model.
We then discard $\sdata$.
We generate internal synthetic datasets of a scale similar to the real training data used for the pre-trained base models: $n_{\rm s}=100$k synthetic samples for CIFAR-10 and FFHQ-64 and $n_{\rm s}=1.5$M synthetic samples for both ImageNet resolutions.

To estimate the distance $\mathrm{dist}(\mathcal{G}, p_{\rm r})$ between the synthetic data distribution $\mathcal{G}$ and the real data  distribution $p_{\rm r}$, we use the Fréchet Inception Distance (FID) \citep{fid}. 
For all generative models and datasets, we generate 50k samples for evaluation, unless stated otherwise. 
Unless specified, we quote the paper-reported metrics for the baseline methods in our comparisons.

\textbf{Quantitative Results.} 
To demonstrate that \proposedMethod{} achieves self-improvement, we need to show that the \proposedMethod{} diffusion model produced by Algorithm~\ref{alg:sims} outperforms the base model.
%
In Figure~\ref{fig:self-improve}, we plot the FID between the \proposedMethod{} model and the real data distribution as a function of the guidance strength parameter $\omega$ and the training budget $\mathcal{B}$
as measured by the number of million-images-seen (Mi) during fine tuning of the auxiliary model.
%
%
In the top row, $\omega=0$ corresponds to no guidance, which establishes the FID attained by the base model. 
%
The key takeaway from Figure~\ref{fig:self-improve} is that, across all four datasets, even a small negative guidance $\omega$ and a small amount of fine-tuning (small Mi) results in a \proposedMethod{} model that outperforms the base model.
Moreover, for properly tuned guidance and training budget, the self-improvement can be substantial:
for CIFAR-10, FFHQ-64, ImageNet-64, and ImageNet-512, \proposedMethod{} yields a relative FID self-improvement of $32.5\%$, $56.9\%$, $41.8\%$, and $32.4\%$, respectively.


\textbf{Qualitative Results.} Figure~\ref{fig:mainfigure} presents example images generated by 
the pre-trained EDM2-S base model (top row), 
the base model after fine-tuning for 102Mi with 1.5M images synthesized from the base model (middle-row), 
and \proposedMethod{} using the same $1.5$M synthetic samples from the base model with a guidance strength of $\omega=0.7$. 
For all three models, we start with the same initial latent vectors. 
We see that \proposedMethod{} qualitatively improves the generated samples in each case. 
Appendices~\ref{appendix:cifar10}--\ref{appendix:ImageNet512} 
present additional qualitative comparisons for the CIFAR10, FFHQ-64, ImageNet-64, and ImageNet-512 datasets.

\begin{table}[t]
\centering
\caption{\small
{\bf \proposedMethod{} attains state-of-the-art image generation performance.}
Image generation performance comparison between \proposedMethod{} and image generation baselines on the CIFAR-10, FFHQ-64, ImageNet-64, and ImageNet-512 datasets.
\proposedMethod{} consistently improves upon the base models EDM-VP and EDM-S. 
Indeed, \proposedMethod{} establishes the new state-of-the-art FID for CIFAR-10 and ImageNet-64 (bold). 
We also compare the number of function evaluations (NFE) required for inference and the number of parameters (Million parameters, Mparams) for each model.
}
\label{tab:comparison_sota}
\begin{minipage}{.48\textwidth}
    \resizebox{\textwidth}{!}{%
\begin{tabular}{lrrr}
\multicolumn{4}{c}{\textbf{CIFAR-10 $32\times 32$ (Unconditional)}}             \\ \hline
Model              & \multicolumn{1}{l}{FID $\downarrow$} & \multicolumn{1}{l}{NFE $\downarrow$} & \multicolumn{1}{l}{Mparams} \\ \hline
DDPM \citep{ho2020denoising}                       & 3.17 & 1000 & -   \\
StyleGAN2-ADA  \citep{karras2020training}        & 2.92 & 1    & -   \\
LSGM \citep{vahdat2021score}                    & 2.10 & 138  & -   \\
NCSN++ \citep{song2020score}                    & 2.20 & 2000 &  -   \\
GDD Distill. \citep{zheng2024diffusion}      & 1.66 & 1    & -   \\
GDD-I Distill. \citep{zheng2024diffusion}    & 1.54 & 1    & -   \\ 
EDM-VP \citep{karras2022elucidating}         & 1.97 & 35   & 280 \\
EDM-G++  \citep{kim2022refining}      & 1.77 & 35   &  -  \\
LSGM-G++ \citep{kim2022refining}                               & 1.94 & 138  & -   \\ \hline
EDM-VP + \proposedMethod{} (Ours)                         & 1.41 & 70   & 560 \\
EDM-VP + \proposedMethod{} + ST (Ours) & \textbf{1.33} & 70               & 560                            \\ \hline
\end{tabular}%
} \vspace{.8em} \\
\resizebox{\textwidth}{!}{%
\begin{tabular}{@{}lrrr@{}}
\multicolumn{4}{c}{\textbf{FFHQ $64\times 64$}}                      \\ \toprule
Model & \multicolumn{1}{l}{FID $\downarrow$} & \multicolumn{1}{l}{NFE $\downarrow$} & \multicolumn{1}{l}{Mparams} \\ \midrule
EDM-VE \citep{karras2022elucidating}      & 2.53 & 79  & 280 \\
EDM-VP \citep{karras2022elucidating}      & 2.39 & 79  & 280 \\
EDM-G++  \citep{kim2022refining}      & 1.98 & 71   &  -  \\
GDD Distill. \citep{zheng2024diffusion}   & 1.08 & 1   & -   \\
GDD-I Distill. \citep{zheng2024diffusion} & \textbf{0.85} & 1   & -   \\ \midrule
EDM-VP + \proposedMethod{} (Ours)                      & 1.04 & 158 & 560 \\
EDM-VP + \proposedMethod{} + ST (Ours)                        & 1.03 & 158 & 560 \\ \bottomrule
\end{tabular}%
}
\vspace{1.3em}
\end{minipage}%
\begin{minipage}{.04\textwidth}
    \hspace{1em}
\end{minipage}%
\begin{minipage}{.48\textwidth}
    \resizebox{\textwidth}{!}{%
\begin{tabular}{@{}lrrr@{}}
\multicolumn{4}{c}{\textbf{ImageNet $64\times 64$}}               \\ \midrule
Model &
  \multicolumn{1}{l}{FID $\downarrow$} &
  \multicolumn{1}{l}{NFE $\downarrow$} &
  \multicolumn{1}{l}{Mparams} \\ \midrule
ADM \citep{dhariwal2021diffusion}  & 2.07 & 250 & - \\
StyleGAN-XL \citep{sauer2022stylegan} & 1.51 & 1 & -\\
RIN  \citep{jabri2023rin}   & 1.23          & 1000 & 280  \\
EDM2-S \citep{karras2024analyzing} & 1.58          & 63   & 280  \\
EDM2-M                             & 1.43          & 63   & 498  \\
EDM2-L                             & 1.33          & 63   & 777  \\
EDM2-XL                            & 1.33          & 63   & 1119 \\
AutoGuidance-S \citep{karras2024guiding} & 1.01          & 126  & 560  \\
GDD-I Distill.  \citep{zheng2024diffusion} & 1.21          & 1    & -    \\ \midrule
EDM2-S + \proposedMethod{} (Ours)               & \textbf{0.92} & 126  & 560  \\ \bottomrule
\end{tabular}%
} \vspace{.5em} \\
\resizebox{\textwidth}{!}{%
\begin{tabular}{lrrr}
\multicolumn{4}{c}{\textbf{ImageNet $512\times 512$ }}                  \\ \hline
Model & \multicolumn{1}{l}{FID $\downarrow$} & \multicolumn{1}{l}{NFE $\downarrow$} & \multicolumn{1}{l}{Mparams} \\ \hline
ADM-G \citep{dhariwal2021diffusion}       & 7.72 & 250  & -    \\
StyleGAN-XL \citep{sauer2022stylegan}     & 2.41 & 1    & -    \\
RIN \citep{jabri2023rin}                  & 3.95 & 1000 & 320  \\
EDM2-S \citep{karras2024analyzing}        & 2.56 & 63   & 280  \\
EDM2-M                                    & 2.25 & 63   & 498  \\
EDM2-L                                    & 2.06 & 63   & 777  \\
EDM2-XL                                   & 1.96 & 63   & 1119 \\
EDM2-XXL                                  & 1.91 & 63   & 1523 \\
AutoGuidance-S \citep{karras2024guiding}  & 1.34 & 126  & 560  \\
AutoGuidance-XL \citep{karras2024guiding} & \textbf{1.25} & 126  & 2236 \\ \hline
EDM2-S + \proposedMethod{} (Ours)                      & 1.73 & 126  & 560  \\ \hline
\end{tabular}%
}
\end{minipage}%
\end{table}

\textbf{Baseline Comparison.} 
Table~\ref{tab:comparison_sota} compares the results obtained by \proposedMethod{} with several standard diffusion based image generation baselines, including  
ADM \citep{dhariwal2021diffusion} optionally used with classifier guidance (ADM-G), 
RIN \citep{jabri2023rin}, 
EDM2-\{S,M,L,XL\} \citep{karras2024analyzing}, DDPM \citep{ho2020denoising}, 
EDM-VP \citep{karras2022elucidating}, 
NCSN++ with improved sampling \citep{song2020score}, 
latent score based model \citep{vahdat2021score}. We also compare with generative adversarial networks (GANs) such as 
StyleGAN-XL \citep{sauer2022stylegan} and StyleGAN-2-ADA \citep{karras2020training}. Additionally, we compare with methods that similar to \proposedMethod{}, improve the performance of a base model, such as  
the distilled single step diffusion models GDD and GDD-I \citep{zheng2024diffusion}, 
discriminator guided models EDM-G++ and LSGM-G++ \citep{kim2022refining}, 
and the EDM2-\{S,XL\} models guided by Autoguidance \citep{karras2024guiding}. 
Note that, for all the aforementioned methods, we present their paper-reported metrics in the table. 
For ImageNet-64 \proposedMethod{} with EDM2-S and for CIFAR-10 \proposedMethod{} with EDM-VP outperforms all of the baseline methods and reaches the new state-of-the-art FIDs of 0.92 and 1.33, respectively, representing a relative improvement of $8.9\%$ and $13.6\%$ over the closest baseline methods, Autoguidance-S and GDD-I. 

Here are two highlights from Table~\ref{tab:comparison_sota}.
First, EDM2-S equipped with \proposedMethod{} surpasses the performance of EDM2-XL by a significant margin for both ImageNet-64 and ImageNet-512, demonstrating that scaling the number of parameters cannot match the performance obtained by training an auxiliary model with synthetic data. 
Second, \proposedMethod{} outperforms discriminator guidance (EDM-G++ and LSGM-G++) by a significant margin for both CIFAR-10 and FFHQ-64, demonstrating that reducing the probability under the synthetic data distribution at each denoising step outperforms increasing the realism score via a discriminator. 
For ImageNet-512, while EDM2-S with \proposedMethod{} outperforms EDM2-S, \proposedMethod{} is outperformed by Autoguidance.

\subsection{MAD Prevention using \proposedMethod{}} 
\label{sec:MPresults}

A fundamental assumption in training a generative model is that the training dataset $\data$ consists exclusively of data that aligns with the ground-truth target distribution. 
When synthetic data generated by previous models is na\"{i}vely included in $\data$ in a self-consuming loop, the the supposed ``ground-truth'' distribution becomes increasingly distorted and ultimately goes MAD. 
In this section we study the abilities of \proposedMethod{} to mitigate and even prevent MADness.

\subsubsection{Two dimensional Gaussian Data in a Synthetic Augmentation Loop}

We now use a simple low-dimensional experiment to demonstrate the effectiveness of \proposedMethod{} in \textit{preventing} the negative impacts of synthetic data training that can lead to MADness. 
Recall from Section~\ref{sec:back} that demonstrating that \proposedMethod{} prevents MAD for a sequence of models $(\mathcal{G}^t)_{t \in \mathbb{N}}$ in a self-consuming loop requires showing that $\mathbb{E}[\mathrm{dist}(\mathcal{G}^\infty, p_{\rm r})] \leq \mathbb{E}[\mathrm{dist}(\mathcal{G}^1, p_{\rm r})]$.

\textbf{Experimental Setup.} We start with the task of learning a simple two-dimensional Gaussian distribution $p_{\rm r} = \mathcal{N}(\vmu, \mSigma)$ with mean $\vmu = [0,0]^{\top}$ and covariance $\mSigma = [2,1;1,2]$ using a DDPM diffusion model \cite{ho2020denoising,ddpmcode}.
We sample a real dataset $\data_{\rm r}$ of size $|\data_{\rm r}| = 1000$ from $\mathcal{N}(\vmu, \mSigma)$ and train the base model $\mathcal{G}^1 = \mathcal{A}(\data_{\rm r})$. 
We then form a synthetic augmentation loop, where for generation $t$ of the loop, $\mathcal{G}^t = \mathcal{A}(\data_{\rm r} \cup \data_{\rm s}^{t-1})$, where $\data_{\rm s}^{t-1}$ is synthetic data generated from the previous generation model $\mathcal{G}^{t-1}$. 
We quantify the performance of the models in terms of the Wasserstein distance $  \mathrm{dist}(\cdot, \cdot) $ between the synthetic and real data distributions $\mathbb{E}[ \mathrm{dist}(\mathcal{G}^t,p_{\rm r}) ]$.

We compare two different training approaches:
\begin{itemize}
    \item \textbf{Standard training}, where we train the generation-$t$ model on the dataset $\data^t = \data_{\rm r}\cup\data_{\rm s}^{t-1}$ in which the real data is {\em polluted} with synthetic data from the previous generation.    
    
    \item \textbf{\proposedMethod{}}, where we train the generation-$t$ base model on the polluted dataset $\data^t$.
\end{itemize}

For both approaches, we trained the base model for 100 epochs on $\data_{\rm r}$. 
For \proposedMethod{}, we obtained the auxiliary model at generation $t$ by fine-tuning the base model for $50$ epochs using $n_{\rm s}=|\sdata| = 2000$  data points synthesized from the base model. 
We calculated expectations over $1000$ independent runs, with each run starting with a new real dataset $\data_{\rm r}$ drawn from $p_{\rm r}$ and continuing the synthetic augmentation loop for $100$ generations.
When there is no guidance ($\omega = 0$),  standard training and \proposedMethod{} coincide and produce identical models.

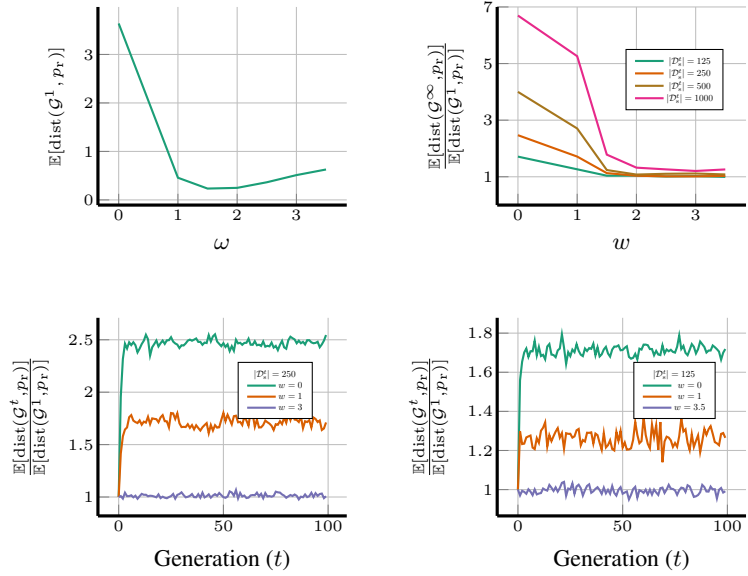
\begin{figure}[t]

\centering
\begin{tikzpicture}
\centering    

\pgfplotsset{/pgfplots/group/every plot/.append style = {
    very thick
}};
    \centering
    \begin{groupplot}[group style = {group size = 2 by 2, horizontal sep = 20mm, vertical sep = 15mm}, width = 0.35\linewidth]

     \nextgroupplot[
        ylabel={\tiny $\mathbb{E}[ \mathrm{dist}(\mathcal{G}^1,p_{\rm r}) ] $},
        xlabel={\small $\omega$},
        axis x line*=bottom,
        axis y line*=left,
        grid, legend style = {at={(0,0)}, nodes={scale=0.35, transform shape}, column sep = 0pt, legend to name = legend1, text=black, cells={align=left},}]

        
        \addplot[colorA, thick]
        table [
            x index=0,
            y index=1,
            col sep=comma]
            {csvfiles/gmad/initialfid.csv};
        
        \nextgroupplot[
        ylabel={\small $\frac{\mathbb{E}[ \mathrm{dist}(\mathcal{G}^{\infty},p_{\rm r}) ]}{\mathbb{E}[ \mathrm{dist}(\mathcal{G}^1,p_{\rm r}) ]} $},
        xlabel={\small $w$},
        axis x line*=bottom,
        axis y line*=left,
        ytick={1,3,5,7},
        ymin=0,ymax=7,
        grid, legend style = {at={(0,0)}, nodes={scale=0.35, transform shape}, column sep = 0pt, legend to name = legend3, text=black, cells={align=left},}]

        \addplot[colorA, thick]
        table [
            x index=0,
            y index=1,
            col sep=comma]
            {csvfiles/gmad/finalperc.csv};
        
        \addplot[colorB,  thick]
        table [
            x index=0,
            y index=2,
            col sep=comma]
            {csvfiles/gmad/finalperc.csv};
        
        \addplot[colorG,  thick]
        table [
            x index=0,
            y index=3,
            col sep=comma]
            {csvfiles/gmad/finalperc.csv};

        \addplot[colorD,  thick]
        table [
            x index=0,
            y index=4,
            col sep=comma]
            {csvfiles/gmad/finalperc.csv};

        \addlegendentryexpanded{$|\data_{\rm s}^t| = 125$};
        \addlegendentryexpanded{$|\data_{\rm s}^t| = 250$};
        \addlegendentryexpanded{$|\data_{\rm s}^t| = 500$};
        \addlegendentryexpanded{$|\data_{\rm s}^t| = 1000$};
        
    \nextgroupplot[
        ylabel={\footnotesize $  \frac{\mathbb{E}[ \mathrm{dist}(\mathcal{G}^t,p_{\rm r}) ]}{\mathbb{E}[ \mathrm{dist}(\mathcal{G}^1,p_{\rm r}) ]} $},
        xlabel={\small Generation ($t$)},
        axis x line*=bottom,
        axis y line*=left,
        grid, legend style = {at={(0,0)}, nodes={scale=0.35, transform shape}, column sep = 0pt, legend to name = legend1, text=black, cells={align=left},}]

        \addlegendimage{empty legend}
        \addlegendentry{\hspace{-1.2cm}$|\data_{\rm s}^t| = 250$}
        
        \addplot[colorA, thick]
        table [
            x index=0,
            y index=1,
            col sep=comma]
            {csvfiles/gmad/allgenNs250.csv};
        
        \addplot[colorB, thick]
        table [
            x index=0,
            y index=2,
            col sep=comma]
            {csvfiles/gmad/allgenNs250.csv};
        
        \addplot[colorC, thick]
        table [
            x index=0,
            y index=3,
            col sep=comma]
            {csvfiles/gmad/allgenNs250.csv};

        \addlegendentryexpanded{$w = 0$};
        \addlegendentryexpanded{$w = 1$};
        \addlegendentryexpanded{$w = 3$};

    \nextgroupplot[
        ylabel={\small $\frac{\mathbb{E}[ \mathrm{dist}(\mathcal{G}^t,p_{\rm r}) ]}{\mathbb{E}[ \mathrm{dist}(\mathcal{G}^1,p_{\rm r}) ]} $},
        xlabel={\small Generation ($t$)},
        axis x line*=bottom,
        axis y line*=left,
        grid, legend style = {at={(0,0)}, nodes={scale=0.35, transform shape}, column sep = 0pt, legend to name = legend2, text=black, cells={align=left},}]

        \addlegendimage{empty legend}
        \addlegendentry{\hspace{-1.2cm}$|\data_{\rm s}^t| = 125$}
        
        \addplot[colorA, thick]
        table [
            x index=0,
            y index=1,
            col sep=comma]
            {csvfiles/gmad/allgenNs125.csv};
        
        \addplot[colorB, thick]
        table [
            x index=0,
            y index=2,
            col sep=comma]
            {csvfiles/gmad/allgenNs125.csv};
        
        \addplot[colorC, thick]
        table [
            x index=0,
            y index=3,
            col sep=comma]
            {csvfiles/gmad/allgenNs125.csv};

        \addlegendentryexpanded{$w = 0$};
        \addlegendentryexpanded{$w = 1$};
        \addlegendentryexpanded{$w = 3.5$};

\end{groupplot}

\node at ($(group c2r1) + (20pt,10pt)$) {\pgfplotslegendfromname{legend3}}; 
\node at ($(group c2r2) + (20pt,10pt)$) {\pgfplotslegendfromname{legend2}}; 
\node at ($(group c1r2) + (20pt,10pt)$) {\pgfplotslegendfromname{legend1}};

\end{tikzpicture}

\caption{\small \textbf{\proposedMethod{} simultaneously self-improves and prevents MADness in the synthetic augmentation self-consuming loop.} 
We compare standard synthetic augmentation training \citep{alemohammad2023arxiv, alemohammad2024selfconsuming} to \proposedMethod{} training in a synthetic augmentation loop across 100 generations for two-dimensional Gaussian data.
Standard training corresponds to guidance $\omega=0$ in all cases.
At top left, we confirm \proposedMethod{}'s {\em self-improvement} by noting that, for a wide range of $\omega$, the expected Wasserstein distance $\mathbb{E}[ \mathrm{dist}(\mathcal{G}^1,p_{\rm r}) ]$ between the first generation model $\mathcal{G}^1 = \mathcal{A}(\data_{\rm r})$ and the real data distribution drops.
At the bottom, we confirm that \proposedMethod{} can act a {\em prophylactic for MADness}.
We plot $ \frac{\mathbb{E}[ \mathrm{dist}(\mathcal{G}^t,p_{\rm r}) ]}{\mathbb{E}[ \mathrm{dist}(\mathcal{G}^1,p_{\rm r}) ]}$, the ratio of the expected Wasserstein Distance at generation $t$ to that at generation 1 for $|\data_{\rm s}^t|=250$ and 125. 
The green/orange/purple curves correspond to weak MADness mitigation/strong MADness mitigation/MADness prevention.
At top right, we plot the normalized expected Wasserstein distance at convergence as a function of $\omega$ for four different synthetic data sizes $|\data_{\rm s}^t|$. 
A guidance parameter of $\omega\approx 3$ results in either strong MADness mitigation or complete MADness prevention. 
} 
\label{fig:G-mad}

\end{figure}

\textbf{Results.} 
First, we confirm \proposedMethod{}'s {\em self-improvement}.
Figure \ref{fig:G-mad} top left plots the expected Wasserstein distance $\mathbb{E}[ \mathrm{dist}(\mathcal{G}^1,p_{\rm r}) ]$ for the first generation model $\mathcal{G}^1 = \mathcal{A}(\data_{\rm r})$ for various values of $\omega$ in \proposedMethod{}.
We see clearly that \proposedMethod{} has exploited its self-synthesized data to self-improve over the base model. trained on purely real data (there is no synthetic data pollution in generation 1).

Next, we confirm that \proposedMethod{} can act a {\em prophylactic against MADness}.
In Figure~\ref{fig:G-mad} bottom, we plot $ \frac{\mathbb{E}[ \mathrm{dist}(\mathcal{G}^t,p_{\rm r}) ]}{\mathbb{E}[ \mathrm{dist}(\mathcal{G}^1,p_{\rm r}) ]}$, the ratio of the expected Wasserstein Distance at generation $t$ to that at generation 1, over 100 synthetic augmentation loop generations for two synthetic dataset sizes: 
$|\data_{\rm s}|=250$ and 125. 
With standard training ($\omega = 0$, green curves), we observe that the Wasserstein distance ratio quickly increases to a value much larger than 1, confirming MADness. 
In words, the performance of models that aggregate the real and synthetic data together and use standard training deteriorates with each generation $t$ in the synthetic augmentation loop until it converges to a stable point, consistent with the findings regarding MADness mitigation in \cite{bertrand2023stability, dohmatob2024a, gillman2024selfcorrecting}. 
However, as $\omega$ increases (orange curves), the \proposedMethod{} Wasserstein distance ratio remains closer to 1, meaning that the negative impacts of synthetic training have been reduced.
Moreover, for an optimized $\omega$ (purple curves), the \proposedMethod{} Wasserstein distance ratio does not deviate from 1, meaning that MADness has been completely {\em prevented}.

To gain insight into the convergence limit for different $\omega$, we calculated $\mathbb{E}[ \mathrm{dist}(\mathcal{G}^{\infty},p_{\rm r}) ]$ by averaging $\{ \mathbb{E}[ \mathrm{dist}(\mathcal{G}^t,p_{\rm r}) ] \}_{t = 20}^{100}$ and plot its ratio to $\mathbb{E}[ \mathrm{dist}(\mathcal{G}^{1},p_{\rm r}) ]$ in Figure \ref{fig:G-mad} top right. 
The minimum values of $\frac{\mathbb{E}[ \mathrm{dist}(\mathcal{G}^{\infty},p_{\rm r}) ]}{\mathbb{E}[ \mathrm{dist}(\mathcal{G}^1,p_{\rm r}) ]}$ over different $\omega$ for $|\data_{\rm s}^t| = 125,250,500,1000$ were $ 0.996, 1.013, 1.078, 1.204 $, respectively. 
The corresponding ratios for standard data training were
$1.71, 2.46, 3.99, 6.69$.

These results suggest that \proposedMethod{} features a {\em prophylactic threshold} on the amount of synthetic data pollution, below which MADness prevention is possible but above which only MADness mitigation is possible.
In this particular experiment, that threshold is approximately $|\data_{\rm s}|=250$.
There are interesting parallels between this property and the fresh data threshold of the fresh data self-consuming loop in \citep{alemohammad2023arxiv, alemohammad2024selfconsuming}.
Exploring and characterizing this threshold are interesting avenues for further research.


To summarize, \textit{to the best of our knowledge, \proposedMethod{} is the first synthetic-data learning algorithm that can prevent MAD in a self-consuming loop without injecting external knowledge.}

\subsubsection{Realistic Data in a Synthetic Augmentation Loop}
\label{sec:realistic}

We continue our exploration of self-improvement and MADness prevention using realistic image data from the CIFAR-10 and FFHQ-64 datasets, large-scale diffusion models, and more pragmatic contexts regarding how the synthetic data enters the synthetic augmentation loop.

We compare four different training scenarios. 
The real dataset $\data_{\rm r}$ (either CIFAR-10 or FFHQ-64) is the same in each scenario.
\begin{itemize}

    \item \textbf{First generation, standard training with purely real data}, $\mathcal{G}^1_{\text{ST-I}}$:
    This scenario corresponds to training a primordial model using standard training and exclusively real data $\data_r$. 
    As an archetype of today's lax data curation practices, data synthesized from $\mathcal{G}^1_{\text{ST-I}}$, which we denote by $\data_{\rm p}$, pollutes the ``real'' training data of the last two second-generation models below. 
    
    \item \textbf{Second generation, ideal \proposedMethod{} training with purely real data}, $\mathcal{G}^1_{\text{SIMS-I}}$: 
    This wishful, idealized scenario corresponds to how synthetic data training should be performed: by applying \proposedMethod{} to self-improve the base model 
    $\mathcal{G}^1_{\text{ST-I}}$ that was trained on purely real data. 
    
    \item \textbf{Second generation, standard training with polluted real data, $\mathcal{G}^2_{\text{ST-P}}$}: 
    This practical scenario corresponds to training a model using standard training with the {\em polluted} training data comprising the purely real data $\data_{\rm r}$ combined with synthetic data $\data_{\rm p}$ generated by $\mathcal{G}^1_{\text{ST-I}}$.
    We know from \citep{alemohammad2023arxiv, alemohammad2024selfconsuming} that this approach leads to MADness.
   
    \item \textbf{Second generation, \proposedMethod{} training with polluted real data, $\mathcal{G}^2_{\text{SIMS-P}}$}: 
    This practical scenario corresponds to training a model using \proposedMethod{} training with the same polluted training data comprising the purely real data $\data_{\rm r}$ combined with synthetic data $\data_{\rm p}$ generated by $\mathcal{G}^1_{\text{ST-I}}$.
    
\end{itemize}

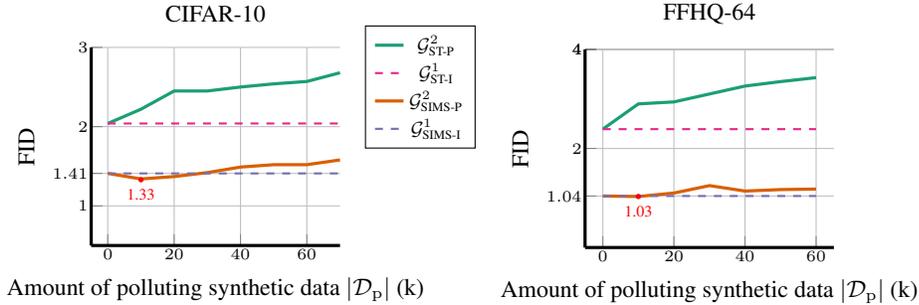
\begin{figure}[t]
    \centering
    \begin{minipage}{0.35\linewidth}
        \centering
        \begin{tikzpicture}

            \begin{axis}[
                title={\footnotesize CIFAR-10},
                ylabel={\footnotesize FID}, 
                xlabel={\footnotesize Amount of polluting synthetic data $|\data_{\rm p}|$ (k)},
                axis x line*=bottom,
                axis y line*=left,
                axis line style = very thick,
                xmin = -5, xmax = 70,
                ymin = 0.5, ymax = 3,
                ytick={1,1.41,2,3},
                grid,
                width=\linewidth,
                legend style={
                at={(1.1,0.8)},
                anchor=west,
                font=\footnotesize},
                legend style={nodes={scale=\legendscale, transform shape}},
                ]

                \addplot[colorA, very thick]
                table [
                    x index=0, 
                    y index=1, 
                    col sep=comma] {csvfiles/synthetic_train/cifar10_synthetic_train_3.csv};
                
                \addplot[colorD,dashed, thick]
                table [
                    x index=0, 
                    y index=4, 
                    col sep=comma] {csvfiles/synthetic_train/cifar10_synthetic_train_3.csv};

                 \addplot[colorB, very thick]
                table [
                    x index=0, 
                    y index=2, 
                    col sep=comma] {csvfiles/synthetic_train/cifar10_synthetic_train_3.csv};

                \addplot[colorC, dashed, thick]
                table [
                    x index=0, 
                    y index=3, 
                    col sep=comma] {csvfiles/synthetic_train/cifar10_synthetic_train_3.csv};

                 \addplot[only marks, mark=*, mark size=.8pt, color=red] coordinates {(10,1.33)};
                \node at (axis cs:10,1.33) [anchor=north] {\tiny \textcolor{red}{1.33}};

                \addlegendentryexpanded{$\mathcal{G}^2_{\text{ST-P}}$};
                \addlegendentryexpanded{$\mathcal{G}^1_{\text{ST-I}}$};
                \addlegendentryexpanded{$\mathcal{G}^2_{\text{SIMS-P}}$};
                \addlegendentryexpanded{$\mathcal{G}^1_{\text{SIMS-I}}$};

            \end{axis}
        \end{tikzpicture}
    \end{minipage}
    \hspace{1.5cm}
        \begin{minipage}{0.35\linewidth}
        \centering
        \begin{tikzpicture}

            \begin{axis}[
                title={\footnotesize FFHQ-64},
                ylabel={\footnotesize FID}, 
                xlabel={\footnotesize Amount of polluting synthetic data $|\data_{\rm p}|$ (k)},
                axis x line*=bottom,
                axis y line*=left,
                axis line style = very thick,
                ymin = 0,ymax = 4,
                xmin=-5,xmax=65,
                ytick={1.04,2,4,4},
                grid,
                width=\linewidth,
                legend style={
                at={(1.1,0.7)},
                anchor=west,
                font=\footnotesize},
                legend style={nodes={scale=\legendscale, transform shape}},
                ]
                    
                \addplot[colorA, very thick]
                table [
                    x index=0, 
                    y index=1, 
                    col sep=comma] {csvfiles/synthetic_train/ffhq64_synthetic_train_2.csv};
                
                \addplot[colorD,dashed, thick]
                table [
                    x index=0, 
                    y index=4, 
                    col sep=comma] {csvfiles/synthetic_train/ffhq64_synthetic_train_2.csv};

                 \addplot[colorB, very thick]
                table [
                    x index=0, 
                    y index=2, 
                    col sep=comma] {csvfiles/synthetic_train/ffhq64_synthetic_train_2.csv};

                \addplot[colorC, dashed, thick]
                table [
                    x index=0, 
                    y index=3, 
                    col sep=comma] {csvfiles/synthetic_train/ffhq64_synthetic_train_2.csv};

                 \addplot[only marks, mark=*, mark size=.8pt, color=red] coordinates {(10,1.03)};
                \node at (axis cs:10,1.03) [anchor=north] {\tiny \textcolor{red}{1.03}};

            \end{axis}
        \end{tikzpicture}
    \end{minipage}
\caption{\small \textbf{\proposedMethod{} acts as a prophylactic against MADness for realistic training datasets polluted with synthetic data.} 
For the CIFAR-10 (50k real images, left) and FFHQ-64 (70k real images, right) datasets, we plot the FID of the four training scenarios from Section~\ref{sec:realistic} as a function of the amount of polluting synthetic data $|\data_{\rm p}|$.
While the modeling performance of standard training is strongly affected by increasing amounts of synthetic data pollution (compare $\mathcal{G}^2_{\text{ST-P}}$ to $\mathcal{G}^2_{\text{ST-I}}$), 
the performance of \proposedMethod{} training is relatively immune (compare $\mathcal{G}^2_{\text{\proposedMethod{}-P}}$ to $\mathcal{G}^2_{\text{\proposedMethod{}-I}}$).}
\label{fig:eliminate_mad}
\end{figure}

\paragraph{Experimental setup.} For $\mathcal{G}^1_{\text{ST-I}}$, we used the EDM-VP models pre-trained on CIFAR-10 and FFHQ-64 from \citep{karras2022elucidating}. 
For CIFAR-10, we trained both $\mathcal{G}^2_{\text{ST-P}}$ and the base model in $\mathcal{G}^2_{\text{SIMS-P}}$ from scratch for $200$Mi. 
%
For FFHQ-64, to reduce computational costs, we fine-tuned $\mathcal{G}^1_{\text{ST-P}}$ and the base model in $\mathcal{G}^2_{\text{SIMS-P}}$ for $100$Mi rather than training from scratch. 
For the training sets $\sdata$ of the auxiliary models in \proposedMethod{}, we generated $|\sdata|=100$k data from the corresponding base models.
For each $|\data_{\rm p}|$, we report the best FID for $\mathcal{G}^2_{\text{SIMS-P}}$ over various values of guidance $\omega$ and training budget $\mathcal{B}$ of the auxiliary model. 
The procedure for $\mathcal{G}^1_{\text{SIMS-I}}$ is identical to the self-improved models for CIFAR-10 and FFHQ-64 in Section \ref{sec:SIresults}, so we re-use those results here.

\paragraph{Results.} 
Figure~\ref{fig:eliminate_mad} plots the FIDs attained by the diffusion models learned by the four training scenarios above for the CIFAR-10 and FFHQ-64 datasets as we vary the amount of synthetic data $|\data_{\rm p}|$ that is polluting the real training dataset.
The same trends occur for both datasets.
First, we see a substantial {\em self-improvement} in modeling performance from $\mathcal{G}^1_{\text{ST-I}}$ to $\mathcal{G}^1_{\text{\proposedMethod{}-I}}$.
Indeed, the drop in FID for CIFAR-10 from $1.41$ (Section \ref{sec:SIresults}) to $1.33$, {\em sets a new state-of-the-art FID benchmark for CIFAR-10 generation.}
Second, we see that increasing amounts of polluting synthetic data $|\data_{\rm p}|$ cause the performance of $\mathcal{G}^1_{\text{ST-P}}$ to diverge from $\mathcal{G}^1_{\text{ST-I}}$.
Third, in contrast to standard training, the performance of \proposedMethod{} training is relatively insensitive to the presence of polluting synthetic data in the base model, which indicates a {\em prophylactic} function against MADness.
More precisely, the plots indicate that, for $|\data_{\rm p}| < 30$k with CIFAR-10 (60\% of $|\data_{\rm r}|$) and $|\data_{\rm p}| < 15$k for FFHQ-64 (20\% of $|\data_{\rm r}|$), \proposedMethod{} not only prevents MADness in the second generation models but also achieves a self-improved FID by somehow exploiting the polluting synthetic data from the previous generation in its training set. 
The reason for this behavior remains an interesting open research question.

Our findings have potential implications for the future of diffusion generative models. 
Previous research has surfaced a ``first mover'' advantage for generative models, whereby large models trained early on real internet data will have a performance edge over later models trained on a mix of real and synthetic data from earlier generation models \citep{alemohammad2023arxiv,alemohammad2024selfconsuming,shumailov2023curse}. 
This advantage for standard training is evident in Figure \ref{fig:eliminate_mad}, where the FID scores of the models degrade as the proportion of synthetic data increases.
In contrast, and somewhat surprisingly, with \proposedMethod{} training, model performance can actually improve when a small amount of synthetic data pollutes the training data.
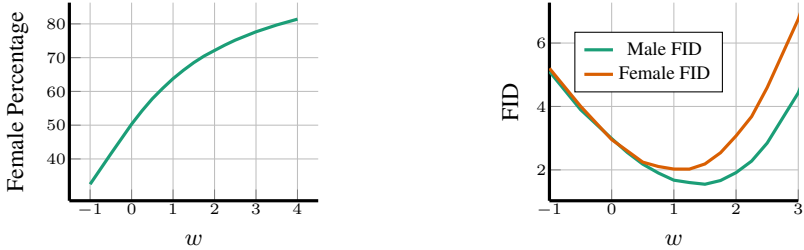
\begin{figure}[t]
    \centering
        \begin{minipage}{0.35\linewidth}
        \centering
        \begin{tikzpicture}

            \begin{axis}[
                ylabel={\small Female Percentage}, 
                xlabel={\small $w$},
                axis x line*=bottom,
                axis y line*=left,
                axis line style = very thick,
                grid,
                xtick={-1,0,1,2,3,4},
                ytick={40,50,60,70,80},
                width=\linewidth,
                legend style={
                at={(0.3,.47)},
                anchor=south west,
                font=\footnotesize},
                legend style={nodes={scale=\legendscale, transform shape}},
                ]
    
                \addplot[colorA, very thick]
                table [
                    x index=0, 
                    y index=1, 
                    col sep=comma] {csvfiles/dist_shift/female_percent.csv};

                
            \end{axis}
        \end{tikzpicture}
    \end{minipage}
    \hspace{1.5cm}
    \begin{minipage}{0.35\linewidth}
        \centering
        \begin{tikzpicture}

            \begin{axis}[
                ylabel={\small FID}, 
                xlabel={\small $w$},
                axis x line*=bottom,
                axis y line*=left,
                axis line style = very thick,
                xmin = -1, xmax = 3,
                grid,
                width=\linewidth,
                legend style={
                at={(0.1,0.7)},
                anchor=west,
                font=\footnotesize},
                legend style={nodes={scale=\legendscale, transform shape}},
                ]

                \addplot[colorA, very thick]
                table [
                    x index=0, 
                    y index=1, 
                    col sep=comma] {csvfiles/dist_shift/male_female_fid.csv};
    
                \addplot[colorB, very thick]
                table [
                    x index=0, 
                    y index=2, 
                    col sep=comma] {csvfiles/dist_shift/male_female_fid.csv};

                \addlegendentryexpanded{Male FID};
                \addlegendentryexpanded{Female FID};

            \end{axis}
        \end{tikzpicture}
    \end{minipage}
    \caption{\small
    \textbf{\proposedMethod{} can simultaneously shift the synthetic distribution to an arbitrary in-domain target distribution while self-improving the quality of generation.} (left) Percentage of female synthetic images for different values of the guidance $\omega$.
    (right) FID of synthetic male and female images with respect to the male and female images in the FFHQ-64 dataset for different guidance levels $\omega$.}
    \label{fig:dis_shift}
\end{figure}

\subsection{Distribution Shifts with \proposedMethod} 
\label{sec:DSresults}


Often, the datasets used for training AI models follow a distribution $p$ that differs from some desired target distribution $\widehat{p}$. 
Consequently, the synthetic data distribution generated by a model will also reflect this discrepancy. 
This technical issue underlies why generative models tend to synthesize biased samples related to demographic factors such as gender and race, which leads to inaccurate representations across these attributes and potentially decreased fairness \citep{friedrich2023fair}.

In this section, we demonstrate that  \proposedMethod{} can align the distribution of its generated images with an arbitrary in-domain target distribution $ \widehat{p} $ that is distinct from the model's training data distribution $p$. 
Simultaneously, we aim to enhance the quality of individual samples. 
By doing so, \proposedMethod{} has the potential to not only self-improve but also mitigate extant biases in a base model by shifting the model distribution towards a different distribution that promotes fairness.

We highlight \proposedMethod{}' abilities for simultaneous self-improvement and distribution shifting with an example of altering group representation frequency using the FFHQ-64 dataset.
This dataset comprises $70$k images of faces varying in gender, age, and race, with an almost equal split of male and female subjects (51\% female and 49\% male). 
The pre-trained EDM-VP model trained on FFHQ-64 in \citep{karras2022elucidating} generate synthetic samples that are 50.3\% perceived female and 49.7\% perceived male \citep{karkkainen2021fairface}. 
This type of generation is arguably fair to both genders, but to demonstrate \proposedMethod{}' ability to adapt to an arbitrary target distribution, our goal is to construct a model that overrepresents females compared to males, changing the percentage to 70\% female and 30\% male.

The synthetic samples we constructed in Section \ref{sec:3} were generated without any intervention in order to match the distribution of the base model's synthetic data. 
Now, we generate samples and use the pre-trained classifier from \citep{karkkainen2021fairface} to label the perceived genders of the generated faces.
Using this information, we construct a synthetic dataset of $140$k images containing 70\% male and 30\% female images. 
Since the score function of the auxiliary model $\vs_{\theta_{\rm s}}(\vx_t, t)$ is used as a negative guidance, the distribution generated by the auxiliary model should be the complement of the target distribution $\widehat{p}$.  
Executing \proposedMethod, we obtain the auxiliary model by fine-tuning the pre-trained diffusion model on FFHQ-64 for 50Mi and then combining the score functions of the base and auxiliary diffusion models with guidance strength $\omega$.


\begin{figure}%
    \centering
    \subfloat[\centering 
    EDM-VP baseline, $\omega = 0$, 50.3\% female]{{\includegraphics[width=6cm]{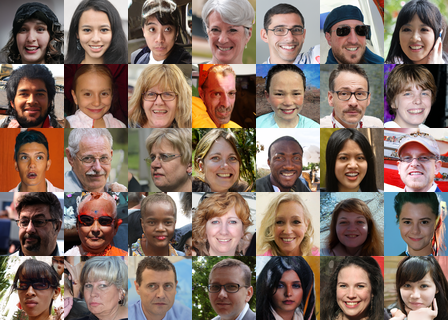} }}%
    \qquad
    \subfloat[\centering \proposedMethod{}, $\omega = 1.5$, 68.5\% female]{{\includegraphics[width=6cm]{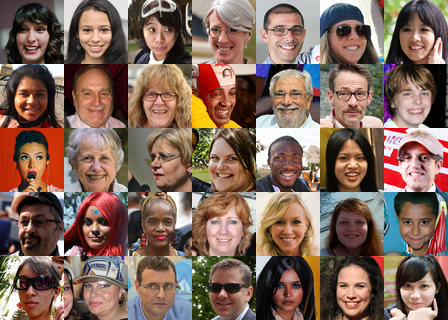} }}
    \caption{\small{\bf Distribution shifting with \proposedMethod{}.} 
    (left) Sample images synthesized from the pre-trained baseline diffusion model EDM-VP from \citep{karras2022elucidating} trained on the FFHQ-64 dataset are approximately 50\% female.
    (right) Sample images synthesized using \proposedMethod{} targeting a distribution shift to approximately 70\% female.
    We used the same seed and randomness for both models to highlight the distribution shift.}
    \label{fig:dis_image}%
\end{figure}

\textbf{Results.} Figure \ref{fig:dis_shift} (left) presents the evidence on distribution shifting.
It plots the percentage of females with respect to the guidance $\omega$. 
For $\omega = -1$, we sample only from the auxiliary model, which has been trained on a synthetic dataset of 70\% males and 30\% females, generating 32\% female images. 
For $\omega = 0$, we sample from the base model and obtain 50\% female images. As $\omega$ increases, the percentage of females increases, reaching approximately 68\% at $\omega = 1.5$.

To asses the quality of image generation, we provide two FID measures: one between synthetic male images and real male images in FFHQ-64, and one between synthetic female images and real female images in FFHQ-64, using 35k synthetic images for each gender. To identify the gender of the synthesized, we again use the pre-trained classifier from \citep{karkkainen2021fairface}.


Figure \ref{fig:dis_shift} (right) presents the evidence on \textit{simultaneous self-improvement}.
It plots the FID scores for the synthesized male and female images. 
The FID follows a bowl-shaped pattern similar to the plots in 
Section \ref{sec:SIresults}.
The minimum FID for male images occurs at $\omega = 1.5$, which coincides with the parameter value that achieves approximately 70\% female generation. 
However, the minimum FID for female images is reached at a slightly lower value of $\omega = 1.25$. This indicates that optimizing both the target distribution shift and the quality of image generation may not necessarily align at the same $\omega$ value.

Figure \ref{fig:dis_image} plots sample synthetic images for the baseline model (left) and the final model that are both distribution-shifted and self-improved (right).

\section{Discussion}
\label{sec:discussion}

In this paper, we have developed {\em Self-IMproving diffusion models with Synthetic data} (\proposedMethod{}), a new training algorithm for generative AI models designed to enhance the performance of diffusion models by using their own synthetic data. 
The key idea is to avoid aggregating the real and synthetic data together into one training dataset --- which can lead to a divergence between the model’s distribution and real-world data (MADness \citep{alemohammad2023arxiv, alemohammad2024selfconsuming,shumailov2023curse}) that diminishes model quality and reinforces biases --- and instead use the synthetic data to provide {\em negative guidance} during the generation process to steer a model’s generative process away from the non-ideal synthetic data manifold and towards the real data distribution. 
\proposedMethod{} provides affirmative responses to questions Q1 and Q2 posed in the Introduction.
In particular, (Q1) \proposedMethod{} establishes new records for realistic synthetic data distribution on two important image datasets (CIFAR-10 and ImageNet-64), while (Q2) to the best of our knowledge, \proposedMethod{} is the first generative AI model that can be iteratively trained on self-generated, synthetic data without going MAD.
As an added bonus, \proposedMethod{} can adjust a diffusion model's synthetic data distribution to match any desired in-domain target distribution, helping mitigate biases and ensure model fairness.

Our experiments have revealed that there is a {\em prophylactic threshold} on the ratio of the amount of synthetic data to the amount of real data that \proposedMethod{} can tolerate before it is incapable of fully preventing MADness.
According to our experiments in Section \ref{sec:MPresults}, this threshold is approximately 60\% for CIFAR-10, 20\% for FFHQ-64, and 25\% for two-dimensional Gaussian data. 
Regardless, above this threshold, \proposedMethod{} continues to mitigate MADness even if it cannot fully prevent it.

As synthetic data continues to proliferate online, na\"{i}ve, unsupervised data collection will someday result in the amount of synthetic data exceeding the prophylactic threshold in standard training datasets and rendering methods like \proposedMethod{} less effective at preventing MADness.
Hence, careful dataset curation using recent advances in watermarking and synthetic data detection \citep{bui2023trustmark, bui2023rosteals, wen2024tree} will be crucial for keeping the amount of synthetic data low enough in tomorrow's training datasets.

The \proposedMethod{} concept and our experimental results point towards a number of interesting open research questions. 
We sketch out four of them here.

First, we conjecture that \proposedMethod{}'s performance might be similar if the auxiliary model differs from the base model but matches the base model's performance across the data domain (e.g., employ two different state-of-the-art diffusion models in Algorithm~\ref{alg:sims}). 
Confirming this could result in new negative-guidance-based training algorithms that are {\em broad spectrum prophylactics} against synthetic data from a range of different generative models.

Second, it seems important to understand why \proposedMethod{} does not just tolerate but capitalizes on synthetic data that is polluting the real training dataset employed by its base model.
This suggests that negative-guidance-based training algorithms have unexplored generalizability properties.

Third, models beyond diffusion models can likely be equipped with self-improvement and MADness prophylactic capabilities.
For instance, we can fine-tune a generative adversarial network (GAN) or variational autoencoder (VAE) base model with its own synthetic data and then design a latent-space sampler like 
Polarity \citep{humayun2022polarity} to reduce the density of generated samples under the synthetic distribution.

Fourth, extending our results on distribution shift, we can envision extending \proposedMethod{} to engage with users through synthetic data and collect feedback to adjust the model's distribution to align with user preferences. 
Interestingly, since the auxiliary model in \proposedMethod{} needs to be trained on a synthetic dataset with complementary characteristics to the target distribution, the synthetic data should be curated based on what users do {\em not} prefer.

\section{Acknowledgement}

SA contributed to this work while he was an intern at Adobe research during summer 2024. SA, AH and RB were supported  by NSF grants CCF-1911094 and IIS-1730574; ONR grants N00014-18-1-2571, N00014-20-1-2534, N00014-23-1-2714, N00014-24-1-2225, and MURI N00014-20-1-2787; AFOSR grant FA9550-22-1-0060; DOE grant DE-SC0020345; DOE grant 140D0423C0076; and a Vannevar Bush Faculty Fellowship, ONR grant N00014-18-1-2047. SA was partially supported by a Ken Kennedy Institute 2023--24 BP Graduate Fellowship.

\bibliography{iclr2024_conference}

\newpage

\appendix
\section{Ablation Studies for \proposedMethod}
\label{appendix:ablation}

In this section, we present ablations on the synthetic dataset size used for training the auxiliary model, FID for different number of function evaluations, and strategies for reducing number of function evalutions during inference.

\begin{figure}[t]
    \hspace{-0.5cm}
    \begin{minipage}{0.30\linewidth}
        \centering
        \begin{tikzpicture}

            \begin{axis}[
                ylabel={\scriptsize FID}, 
                xlabel={\scriptsize $\mathcal{B}$ (Mi)},
                axis x line*=bottom,
                axis y line*=left,
                ymin=0.8,ymax=1.6,
                xmin=0,xmax=200,
                axis line style = very thick,
                grid,
                width=\linewidth,
                legend style={
                at={(0.42,0.25)},
                anchor=west,
                font=\footnotesize},
                legend style={nodes={scale=0.4, transform shape}},
                ]

            \addplot[colorA,very thick]
        table [
            x index=0,
            y index=1,
            col sep=comma]
            {csvfiles/ablation/training_size/imagnet64_small_training_different_training.csv};

            \addplot[colorB,very thick]
        table [
            x index=0,
            y index=1,
            col sep=comma]
            {csvfiles/ablation/training_size/imagnet64_100_different_training.csv};

            \addplot[colorC,very thick]
        table [
            x index=0,
            y index=1,
            col sep=comma]
            {csvfiles/ablation/training_size/imagnet64_big_training_different_training.csv};

             \legend{$|\mathcal{D}_s| = 200k$, $|\mathcal{D}_s| = 750k$ ,  $|\mathcal{D}_s| = 1.5M$  }
                
            \end{axis}
        \end{tikzpicture}
    \end{minipage}
    \hspace{-0.55cm}
    \begin{minipage}{0.01\linewidth}
        \vspace{-0.45cm}
        \rotatebox{90}{ \scriptsize FID}
    \end{minipage}
    \hspace{-0.75cm}
    \begin{minipage}{0.30\linewidth}
        \centering
        \begin{tikzpicture}

            \begin{axis}[
                xlabel={\scriptsize NFE},
                axis x line*=bottom,
                axis y line*=left,
                ymin=0.5,ymax=4,
                xmin=0,xmax=140,
                axis line style = very thick,
                grid,
                width=\linewidth,
                legend style={
                at={(0.45,0.75)},
                anchor=west,
                font=\footnotesize},
                legend style={nodes={scale=0.4, transform shape}},
                ]

             \addplot[colorA, very thick]
        table [
            x index=0,
            y index=1,
            col sep=comma]
            {csvfiles/ablation/num_steps/imagnet64_num_steps_wo_guidance.csv};

            \addplot[colorB, very thick]
        table [
            x index=0,
            y index=1,
            col sep=comma]
            {csvfiles/ablation/num_steps/imagnet64_num_steps_w_guidance.csv};
        
        \legend{w/o Guidance , w/ Guidance }
                
            \end{axis}
        \end{tikzpicture}
    \end{minipage}
    \hspace{-0.45cm}
    \begin{minipage}{0.01\linewidth}
        \vspace{-0.45cm}
        \rotatebox{90}{ \scriptsize FID}
    \end{minipage}
    \hspace{-0.65cm}
    \begin{minipage}{0.30\linewidth}
        \centering
        \begin{tikzpicture}

            \begin{axis}[
                xlabel={\scriptsize $\mathcal{B}$ (Mi)},
                axis x line*=bottom,
                axis y line*=left,
                ymin=0.8,ymax=1.6,
                xmin=0,xmax=200,
                axis line style = very thick,
                grid,
                width=\linewidth,
                legend style={
                at={(0.42,0.25)},
                anchor=west,
                font=\footnotesize},
                legend style={nodes={scale=0.4, transform shape}},
                ]

                \addplot[colorA, very thick]
            table [
                x index=0,
                y index=1,
                col sep=comma]
                {csvfiles/ablation/parameters/imagnet64_100_different_training.csv};
            
            \addplot[colorB, very thick]
            table [
                x index=0,
                y index=1,
                col sep=comma]
                {csvfiles/ablation/parameters/imagnet64_decoder_only_different_training.csv};

             \legend{Full params , Decoder only }
                
            \end{axis}
        \end{tikzpicture}
    \end{minipage}
    \hspace{-0.5cm}
    \begin{minipage}{0.01\linewidth}
    \vspace{-0.45cm}
      \rotatebox{90}{ \scriptsize $t_h$}
    \end{minipage}
    \begin{minipage}{0.21\linewidth}
            \centering
            \includegraphics[width=\linewidth]{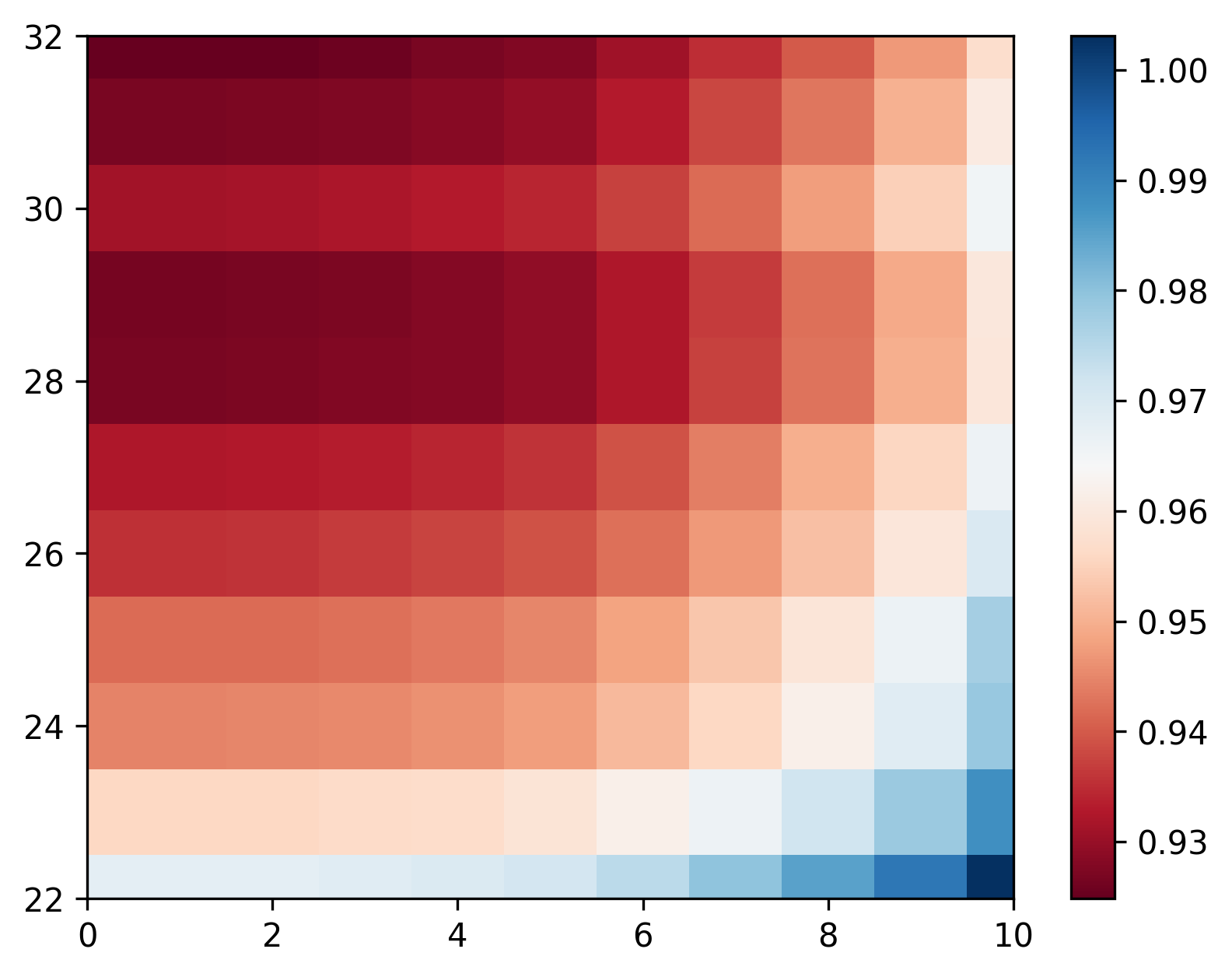}\\
            \hspace{-0.35cm}
            \scriptsize $t_l$
    \end{minipage}
    \hspace{-0.1cm}
    \begin{minipage}{0.01\linewidth}
    \vspace{-0.45cm}
      \rotatebox{90}{ \scriptsize FID}
    \end{minipage}
    \caption{\small \textbf{Left}: training the auxilary model score function $\vs_{\theta_{\rm s}}(\vx,t)$ using synthetic datasets of varying size for ImageNet-64. Increasing synthetic dataset size helps obtain better FID during self-improvement with diminishing returns. 
    \textbf{Middle-left}: FID for different number of function evaluations (NFE).
    \textbf{Middle-right} 
    Reducing the number of learnable parameters during auxiliary model fine-tuning.
    \textbf{Right}
    Changing the guidance interval for \proposedMethod{}. Early and late denoising steps can be ignored with a minimal drop in FID.
    }
    \label{fig:ablations}
\end{figure}
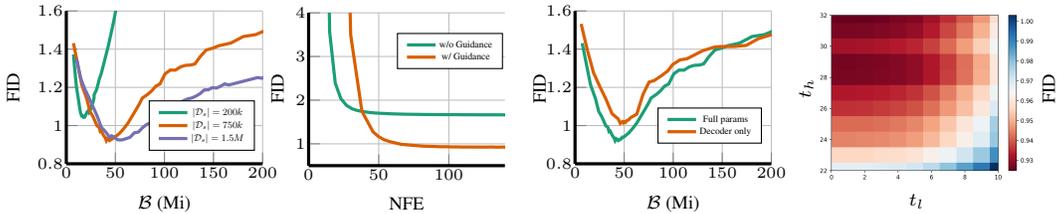

\paragraph{Synthetic dataset size.} For ImageNet-64, we change the dataset size used for training the auxiliary model score function $\vs_{\theta_{\rm s}}(\vx,t)$, and present the FID over training budget. In Figure \ref{fig:ablations} (left), we see that increasing the dataset size allows obtaining better FID. However note that if $|\mathcal{D}_s| \rightarrow \infty$, $\vs_{\theta_{\rm s}}(\vx,t) \rightarrow \vs_{\theta_{\rm r}}(\vx,t)$, i.e., the score functions become identical and negative guidance yields no gain. Therefore increasing the synthetic dataset further to very large numbers may result in an decrease in FID. 

\paragraph{Number of function evaluations.} Number of function evaluations (NFE) refer to the number of times a score function is evaluated during denoising. For ImageNet-64 we compare NFE for the EDM2-S base model with and without \proposedMethod{}. In Figure \ref{fig:ablations} (middle left), we see that naturally, with \proposedMethod{} we need more function evaluations to achieve the lowest FID. At NFE$=40$, FID for both with and without guidance cases are almost equal to $1.70$. For the \proposedMethod{} we use a guidance strength of $\omega=0.9$ and the best FID auxiliary model trained upto $56$ Mi seen during training.

\paragraph{Reducing number of function evaluations.} For a fixed denoising step, \proposedMethod{} uses twice the number of function evaluations (NFE) compared to the baseline method without any guidance. This results in doubling the inference time computation. We propose two strategies to reduce the NFE overhead.

The EDM model architecture consists of an encoder and a decoder, each responsible for half of the computations for one function evaluation. As illustrated in Figure \ref{fig:ablations} (middle right), during the fine-tuning of the base model, we froze the weights of the encoder and trained only the decoder part. At inference time, the encoder is shared between the base model and the auxiliary model, differing only in the decoder. Consequently, the effective number of function evaluations decreases from 2x to 1.5x. We observe that training only the decoder to obtain the auxiliary model slightly increases the minimum FID from $0.92$ to $1.01$ during fine-tuning while reducing the NFE from 2 to 1.5.

The second strategy involves applying guidance from the auxiliary model for a limited interval. To assess the impact of this guidance at different denoising steps, we compute the FID for \proposedMethod{} with guidance applied to a limited interval $(t_l, t_h)$, rather than the default setting of $(0, 32)$. As shown in Figure~\ref{fig:ablations} (right), guidance is more crucial during the final denoising steps compared to the earlier ones. The results indicate that we can exclude the first 10 steps in the denoising process with only a minimal drop in FID, from 0.93 to 0.96. Utilizing the auxiliary model for guidance over a smaller number of intervals can effectively reduce inference time and costs.




\newpage

\section{CIFAR-10 Synthesized Images}
\label{appendix:cifar10}
\begin{figure}[h]%
    \centering
    \subfloat[\centering \proposedMethod{}: $w = 0.8$, Training budget: $40$ Mi]{{\includegraphics[width=14cm]{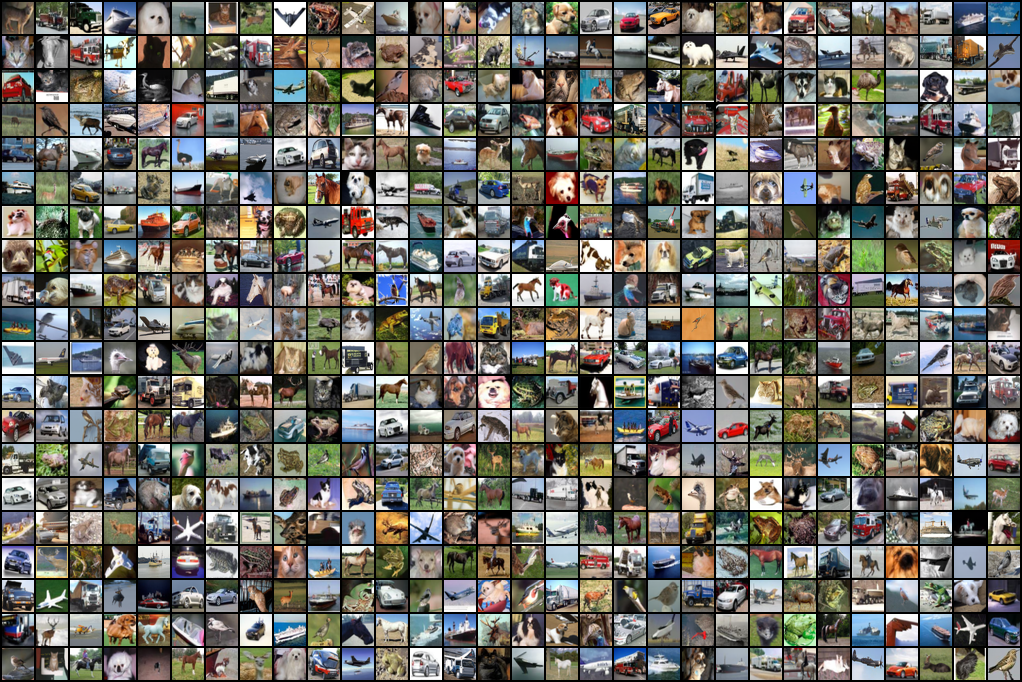} }}%
    \\
    \subfloat[\centering Base Model]{{\includegraphics[width=14cm]{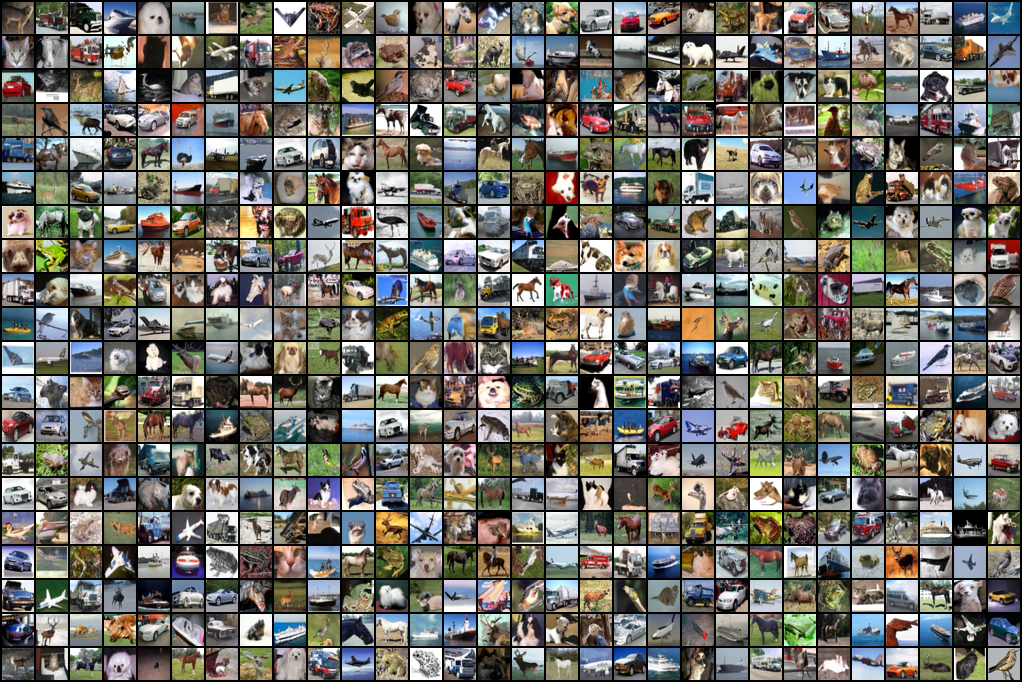} }}%

\end{figure}

\newpage

\section{FFHQ-64 Synthesized Images}
\label{appendix:ffhq}
\begin{figure}[h]%
    \centering
    \subfloat[\centering \proposedMethod{}: $w = 1.5$, Training budget: $34$ Mi]{{\includegraphics[width=14cm]{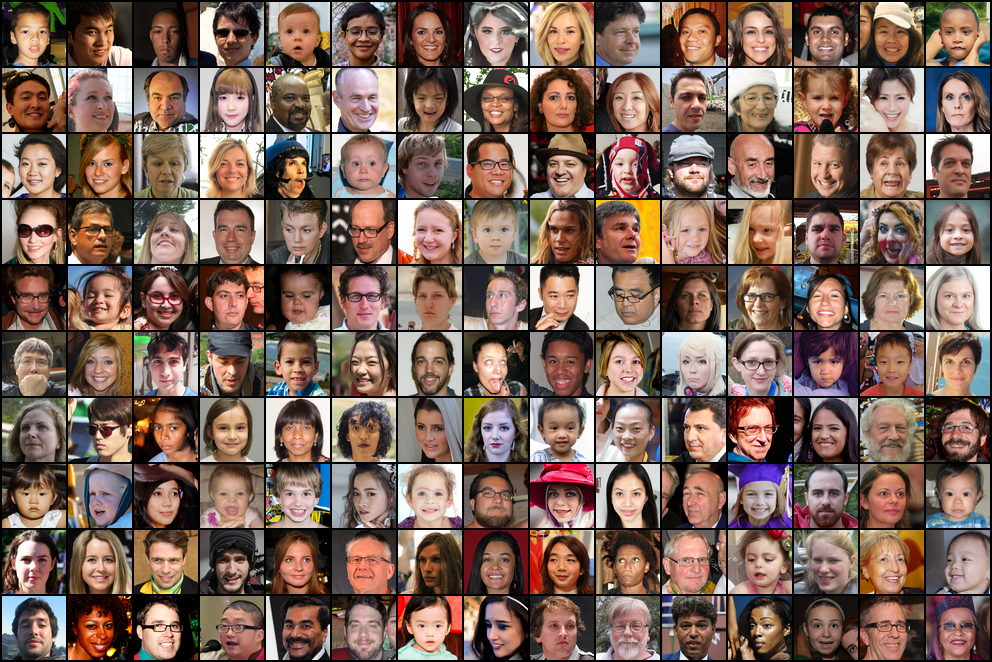} }}%
    \\
    \subfloat[\centering Base Model]{{\includegraphics[width=14cm]{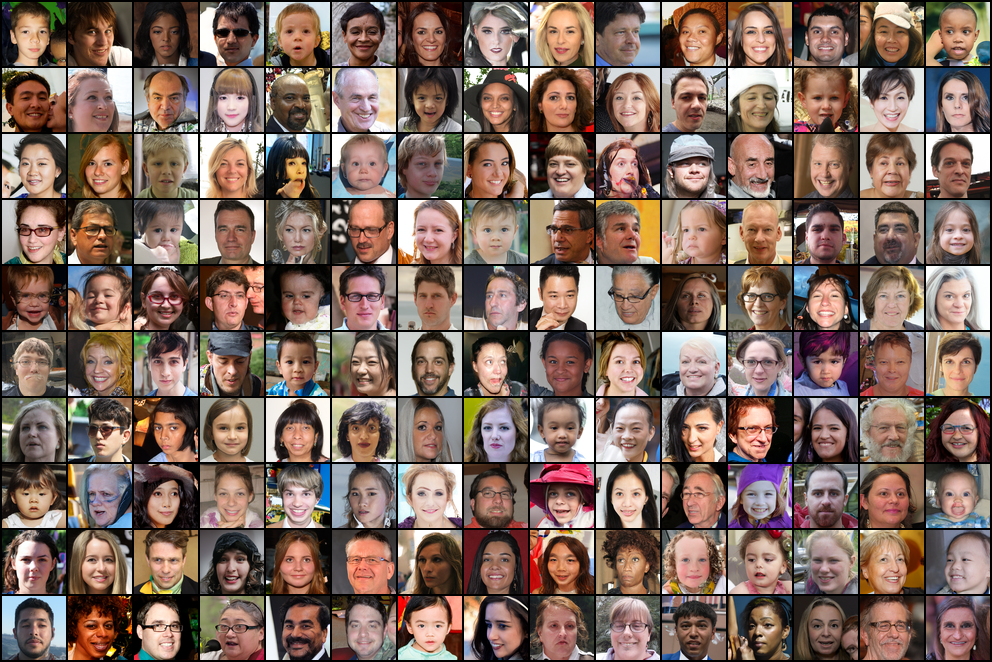} }}%

\end{figure}

\newpage

\section{ImageNet-64 Synthesized Images}
\label{appendix:ImageNet64}
\begin{figure}[h]%
    \centering
    \subfloat[\centering \proposedMethod{}: $w = 0.9$, Training budget: $56$ Mi]{{\includegraphics[width=14cm]{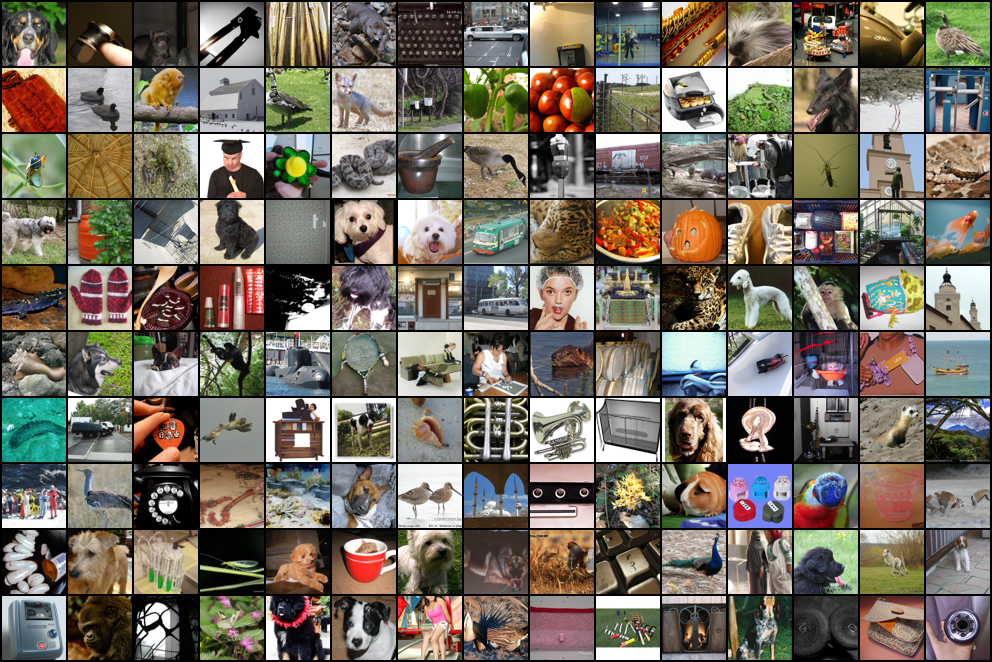} }}%
    \\
    \subfloat[\centering Base Model]{{\includegraphics[width=14cm]{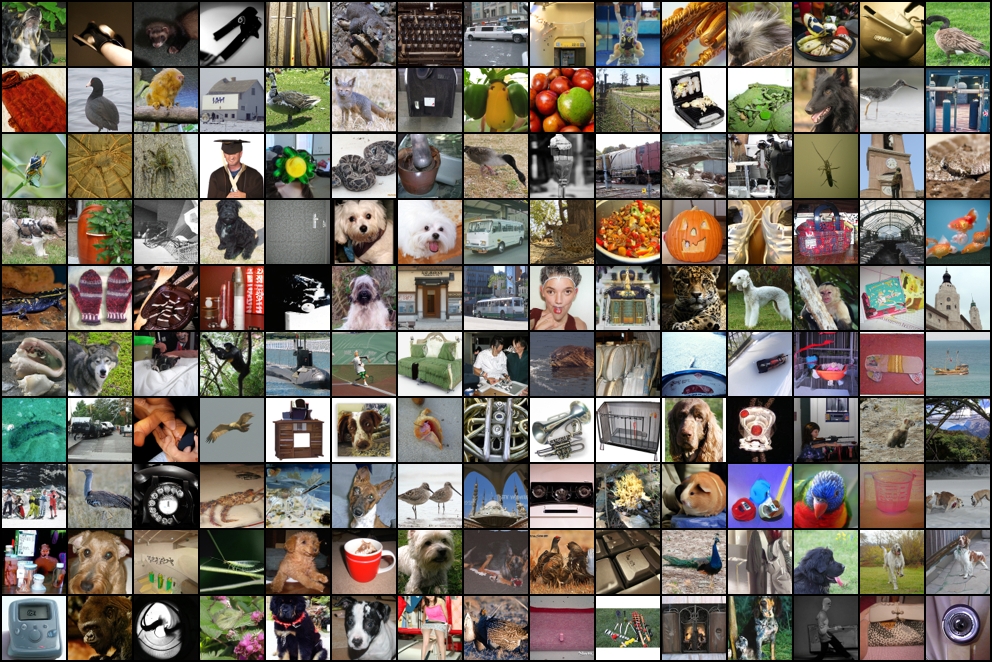} }}%

\end{figure}

\newpage

\section{ImageNet-512 Synthesized Images}
\label{appendix:ImageNet512}

\begin{figure}[h]%
    \centering
    \subfloat[\centering \proposedMethod{}: $w = 0.7$, Training budget: $102$ Mi]{{\includegraphics[width=14cm]{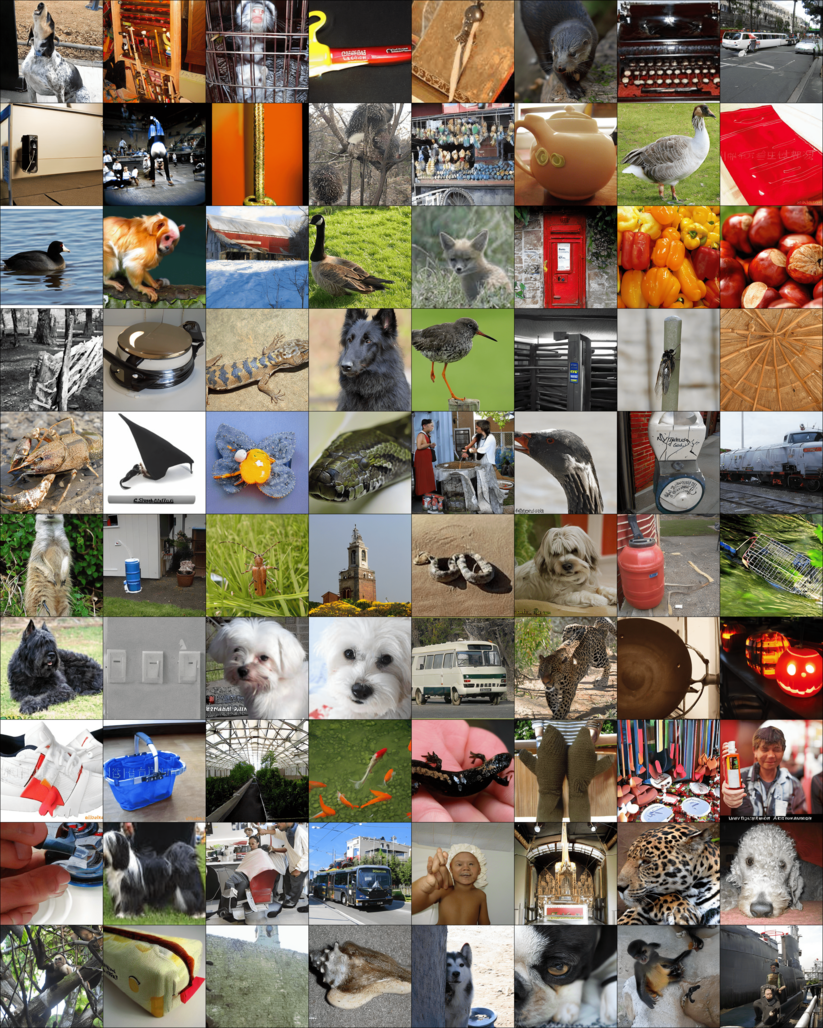} }}%
\end{figure}

\begin{figure}[h]%
    \centering
    \subfloat[\centering Base Model]{{\includegraphics[width=14cm]{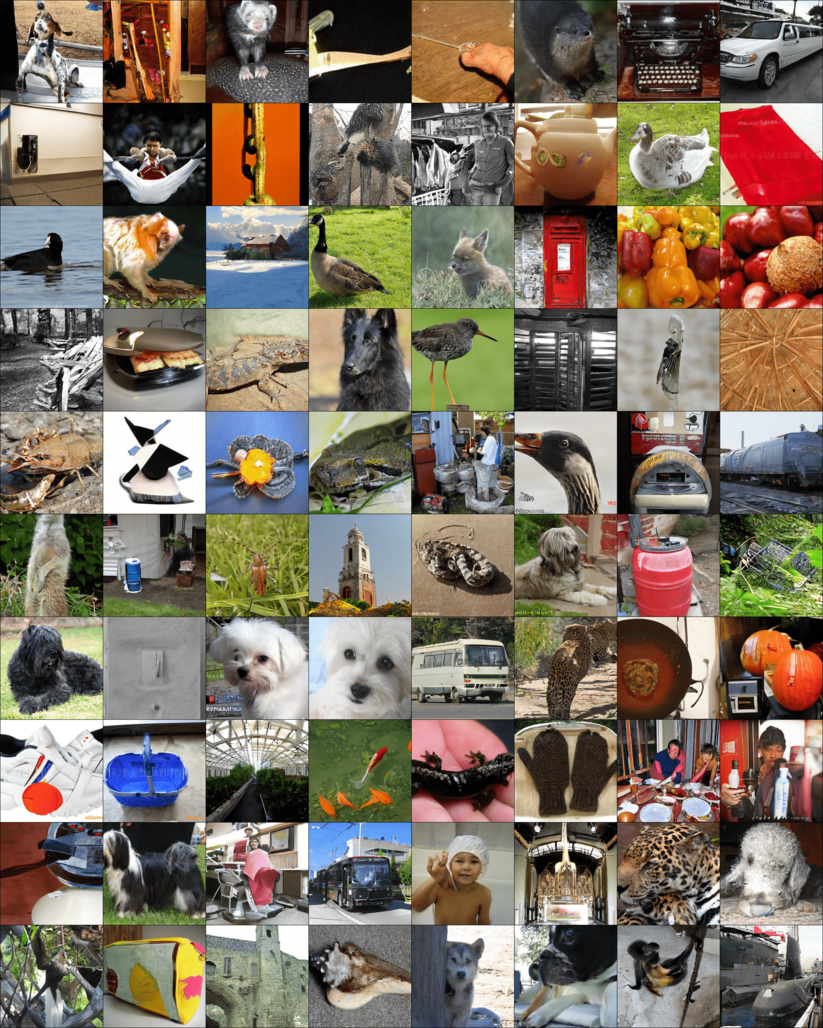} }}%
\end{figure}
\clearpage
\newpage

\section{Standard training}
\label{appendix:ST}

\begin{algorithm}
\caption{\small Standard Training Procedure} \label{alg:st}
\begin{algorithmic}[1]
    \Statex {\bf Input}: Training dataset $\mathcal{D}$
    
    \State \textbf{Train diffusion model}: Use dataset $\data$ to train the diffusion model using standard training, resulting in the score function $\vs_\theta(\vx_t, t)$.
    
    \Statex \textbf{Synthesize}:
    Generate synthetic data from the model using the score function $\vs_{\theta}(\vx_t, t)$.
    
\end{algorithmic}
\end{algorithm}

The procedure of standard training is shown in Algorithm \ref{alg:st}. Compared to \proposedMethod{} (Algorithm \ref{alg:sims}), standard training is essentially the same as using only the base diffusion model's score function to generate synthetic data, which is equivalent to setting $\omega = 0$ in \proposedMethod{}. It's important to note that if you already have a model trained using the standard approach, you can still apply steps 2-4 of \proposedMethod{} to develop a self-improved model.

\end{document}